\pdfoutput=1
\documentclass[11pt]{article}
\usepackage{makecell}

\usepackage{times}
\usepackage{todonotes}
\usepackage{latexsym}
\usepackage{multirow}
\usepackage{booktabs}
\usepackage{soul}
\usepackage{hyperref}
\usepackage[inkscapeformat=png]{svg}
\usepackage[T1]{fontenc}
\usepackage{todonotes}
\usepackage{acl}

\usepackage[utf8]{inputenc}
\usepackage{microtype}
\usepackage{graphicx}
\usepackage{amsmath}
\usepackage{amssymb}
\usepackage{arydshln}

\usepackage[utf8]{inputenc}

\definecolor{ForestGreen}{RGB}{34,139,34}

\usepackage{array} 
\usepackage{colortbl} 
\usepackage{pdflscape} 

\newcolumntype{Y}{>{\centering\arraybackslash}X}

\usepackage{booktabs}
\usepackage{multirow}
\usepackage{graphicx}
\usepackage{subcaption}

\title{\textit{Counterspeech the ultimate shield!} Multi-Conditioned Counterspeech Generation through Attributed Prefix Learning}

\author{Aswini Kumar$^1$,  \textbf{ Anil Bandhakavi}$^2$,  \textbf{Tanmoy Chakraborty}$^1$ \\
$^1${IIT Delhi, India}, $^2${Logically.ai}\\
\small {
\{
\texttt{aswinikumarpadhi1995@gmail.com}, 
\}
}\\
\small
{\tt  anil@logically.ai},
{\tt tanchak@iitd.ac.in}
}

\begin{document}
\maketitle

\begin{abstract}
Counterspeech has proven to be a powerful tool to combat hate speech online. Previous studies have focused on generating counterspeech conditioned only on specific strategies (single attributed). However, a holistic approach considering multiple attributes simultaneously can yield more nuanced and effective responses. Here, we introduce \texttt{HiPPrO}, \textbf{Hi}erarchical \textbf{P}refix learning with \textbf{Pr}eference \textbf{O}ptimization, a novel two-stage framework that utilizes the effectiveness of attribute-specific prefix embedding spaces hierarchically optimized during the counterspeech generation process in the first phase. Thereafter, we incorporate both reference and reward-free preference optimization to generate more constructive counterspeech. Furthermore, we extend \texttt{IntentCONANv2} by annotating all $13,973$ counterspeech instances with emotion labels by five annotators. {\tt HiPPrO} leverages hierarchical prefix optimization to integrate these dual attributes effectively. An extensive evaluation demonstrates that {\tt HiPPrO} achieves a $\sim 38\%$ improvement in strategy conformity and a $\sim 3\%$, $\sim 2\%$, $\sim 3\%$ improvement in Rouge-$1$, Rouge-$2$, and  Rouge-L, respectively, compared to several baseline models. Human evaluations further substantiate the superiority of our approach, highlighting the enhanced relevance and appropriateness of the generated counterspeech. This work underscores the potential of multi-attribute conditioning in advancing the efficacy of counterspeech generation systems.\footnote{\textit{\color{red} Warning: The materials presented in this paper might be disturbing or offensive.}} Our code is available on \href{https://github.com/AswiniNLP/HiPPrO.git}{Github} and dataset is open-sourced on \href{https://huggingface.co/datasets/Aswini123/MultiCONAN}{Hugging-face}. 

\end{abstract}
\section{Introduction}

\begin{figure}[t]
\includegraphics[width=\columnwidth, height=1.1\columnwidth]{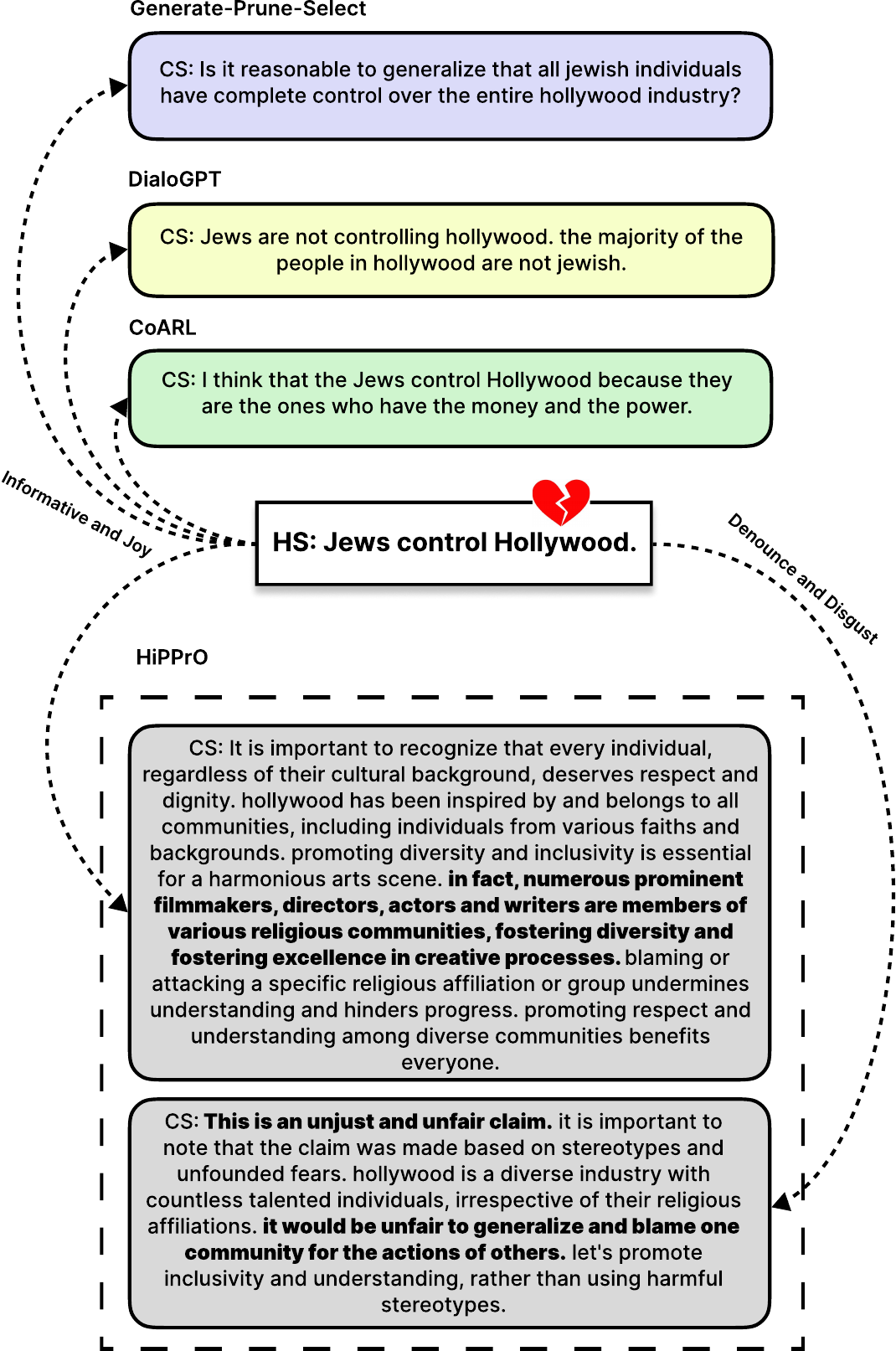}
\caption{An illustration of the output of existing methods in generating effective counterspeech, {\tt HiPPrO} (ours) demonstrates superior performance in producing high-quality, effective, multi-attributed counterspeech for a given hate speech without relying on instructional prompts.}
\label{fig:figure1}
\vspace{-7mm}
\end{figure}

The rise of the Internet has transformed social media platforms into hostile grounds for hateful comments targeting specific communities. Hate speech (HS) \citep{10.1007/978-3-030-75762-5_55,10.1145/3522598.3522601,9458789,masud2023overviewhasocsubtrack2023} carries offensive statements that leverage stereotypes to spread harmful content. In contrast, counterspeech (CS) \cite{benesch2016-considerations-for-succesful, wright2017vectors,singh2024independentfactcheckingorganizationsexhibit} involves constructive responses to counteract hate speech by promoting positive dialogue, thus mitigating online hostility while preserving diverse perspectives \cite{schieb2016governing, chandrasekharan2017you}. As hateful comments surge \cite{leetaru-2019, masud2021hate,masud2022proactively,yadav2024toxbartleveragingtoxicityattributes}, depending only on human-generated counterspeech becomes impractical. In this context, machine learning models appear essential for analyzing and generating counterspeech, offering a promising solution for automating the process. By leveraging this, content moderators can efficiently counter online hate \cite{parker-et-al-2023,garg2023handling, chung-et-al-2023-understanding-counterspeech}. Consequently, several interests \cite{mathew2019thou, qian2019benchmark,fanton2021human, bonaldi-etal-2022-human, hengle2024intentconditioned} have intensified in the development of counterspeech generation methods.

The conceptualization of CS generation has evolved from a simple sequence-to-sequence problem to a more nuanced approach, acknowledging the diverse and context-dependent nature of hate speech \cite{chung2019conan,mathew2019thou,sheng2020towards,parker-et-al-2023,chung-et-al-2023-understanding-counterspeech}. This paradigm shift has fostered the development of advanced generative models incorporating stylistic and condition-guided elements, such as politeness, joyfulness, and detoxification, to create more effective counter-narratives \cite{saha2022countergedi,sheng2020towards}. Recent research has introduced strategy-specific CS generation, where established strategies guide the generation process to combat hate speech \cite{gupta-etal-2023-counterspeeches,hengle2024intentconditioned,benesch-2016-vectors-for-counterspeech}.

\paragraph{\large Motivation:} A significant portion of online hate speech consists of short, abusive statements \cite{benesch2016-considerations-for-succesful}, and CS has shown potential in effectively countering such harmful content. While conventional approaches generate a single response per hate speech instance \cite{zhu-bhat-2021-generate, qian2019benchmark}, recent studies highlight the benefits of tailoring CS to specific attributes for generating more diverse responses \cite{gupta-etal-2023-counterspeeches, hengle2024intentconditioned}. In practical scenarios, hateful comments often include multiple user intentions, necessitating the development of counterspeech that effectively addresses diverse attributes, resulting in more comprehensive and effective responses. For instance, single-attribute approaches typically produce one-dimensional responses by focusing solely on strategy (e.g., being informative) while neglecting the emotional harmony required for persuasive communication. Real-world hate speech frequently needs a range of emotional responses, making it imperative to generate counterspeech that balances factual accuracy with emotional engagement.

The integration of Large Language Models (LLMs) for various text generation tasks has become increasingly popular \cite{yang2024harnessing}, but training these models is resource intensive. To address this, parameter-efficient fine-tuning (PEFT) techniques, such as tunable prefixes \cite{li-liang-2021-prefix}, have gained popularity. These techniques involve adding task-specific continuous vectors (key-value pairs) to transformer layers while keeping the rest of the model unchanged. Recent studies \cite{liu2023pre} showed that these vectors excel in generating conditional text by capturing hidden implied relationships during training. In this work, we propose generating multi-attribute guided counterspeech generation through a hierarchical approach to learning continuous prefix vectors, enabling more varied and contextually relevant responses. We provide an example in Figure \ref{fig:figure1}, where our method, {\tt HiPPrO}, is compared to leading models like Generate-Prune-Select (GPS) \cite{zhu-bhat-2021-generate}, DialoGPT \cite{zhang2020dialogpt}, and COARL \cite{hengle2024intentconditioned}.
While traditional models produce semantically sound responses, they often lack nuance and struggle with complex contextual relationships, as mentioned in \citep{benesch2016-considerations-for-succesful}. Inspired by \citet{10204197}, {\tt HiPPrO} enables to generate more effective and contextually relevant counterspeech by analyzing hate speech and multiple user intentions, resulting in more impactful and persuasive responses.

\paragraph{\large Our Contribution:} This study introduces an advanced pipeline for counterspeech generation, addressing the implicit nature of hate speech with responses aligned to multiple attributes. We focus on four primary strategies-- `positive', `informative', `questioning', and `denouncing.' and five emotion categories --`anger', `disgust', `joy', `sadness', and `surprise'. Furthermore, we curate \texttt{MultiCONAN}, the largest collection of strategy-emo-specific counterspeech, with $13,952$ responses countering $3,487$ instances of hate speech. We propose \texttt{HiPPrO}, a novel two-phase framework that first learns attribute-specific prefix embeddings (key-value pairs) and then applies preference tuning to generate constructive, non-toxic responses. Comprehensive evaluations using automated metrics and human evaluations demonstrate that {\tt HiPPrO} consistently outperforms existing methods in CS generation across various criteria and achieves comparable performance with state-of-the-art LLMs like GPT-3.5 and GPT-4.

\section{Related Work}

The evolution of CS datasets has progressed from crowdsourced collections to expert-curated, strategy-specific compilations \cite{qian2019benchmark,chung2019conan,fanton2021human,gupta-etal-2023-counterspeeches,hengle2024intentconditioned}, while CS generation techniques have advanced from basic sequence-to-sequence models to sophisticated multi-phase pipelines with attribute control \cite{zhu-bhat-2021-generate,saha2019complex}. These advancements have significantly improved the nuance and effectiveness of automated CS in promoting constructive dialogue \cite{benesch2016counterspeech}. Concurrently, parameter-efficient fine-tuning methods like Prefix Tuning \cite{li-liang-2021-prefix} and prompt tuning \cite{lester-etal-2021-power} have emerged, modifying inputs while preserving language model parameters. Recent developments in preference tuning include RLHF's application to instruction-following tasks \cite{ouyang2022training}, Direct Preference Optimization (DPO) \cite{rafailov2023direct}, and a unified approach omitting both reward and reference models \cite{hong2024reference}, addressing challenges in scalability and model sensitivity.

\begin{table*}[]
\centering
\scalebox{1}{
\setlength{\tabcolsep}{1mm} 
\fontsize{7.7pt}{12pt}\selectfont 
\begin{tabular}{| l | c | c | c | c | c | c | c | c | c | c | c | c | c | c | c | c | c | c | c | c | c |}
\hline
\textbf{ }  & \multicolumn{21}{c}{\textbf{Counterspeech strategy and Emotion}}                                                                                                                                                                                                                                                                                                                                                                                                                                                                                                                                                                                                                                                                                                                                                                                                                                                                                                                                                                                                                                                                                                                                                                                                                                                                                                                                                                                                                                                                                                                                                                                                                                                                                                                                                                                 \\ \hline
\textbf{Target} & \multicolumn{1}{c|}{\textbf{\begin{tabular}[c]{@{}c@{}}IN\\ and\\ AN\end{tabular}}} & \multicolumn{1}{c|}{\textbf{\begin{tabular}[c]{@{}c@{}}IN\\ and\\ DI\end{tabular}}} & \multicolumn{1}{c|}{\textbf{\begin{tabular}[c]{@{}c@{}}IN\\ and\\ JO\end{tabular}}} & \multicolumn{1}{c|}{\textbf{\begin{tabular}[c]{@{}c@{}}IN\\ and\\ SA\end{tabular}}} & \multicolumn{1}{c|}{\textbf{\begin{tabular}[c]{@{}c@{}}IN\\ and\\ SU\end{tabular}}} & \multicolumn{1}{c|}{\textbf{\begin{tabular}[c]{@{}c@{}}DE\\ and\\ AN\end{tabular}}} & \multicolumn{1}{c|}{\textbf{\begin{tabular}[c]{@{}c@{}}DE\\ and\\ DI\end{tabular}}} & \multicolumn{1}{c|}{\textbf{\begin{tabular}[c]{@{}c@{}}DE\\ and\\ JO\end{tabular}}} & \multicolumn{1}{c|}{\textbf{\begin{tabular}[c]{@{}c@{}}DE\\ and\\ SA\end{tabular}}} & \multicolumn{1}{c|}{\textbf{\begin{tabular}[c]{@{}c@{}}DE\\ and\\ SU\end{tabular}}} & \multicolumn{1}{c|}{\textbf{\begin{tabular}[c]{@{}c@{}}PO\\ and\\ AN\end{tabular}}} & \multicolumn{1}{c|}{\textbf{\begin{tabular}[c]{@{}c@{}}PO\\ and\\ DI\end{tabular}}} & \multicolumn{1}{c|}{\textbf{\begin{tabular}[c]{@{}c@{}}PO\\ and\\ JO\end{tabular}}} & \multicolumn{1}{c|}{\textbf{\begin{tabular}[c]{@{}c@{}}PO\\ and\\ SA\end{tabular}}} & \multicolumn{1}{c|}{\textbf{\begin{tabular}[c]{@{}c@{}}PO\\ and\\ SU\end{tabular}}} & \multicolumn{1}{c|}{\textbf{\begin{tabular}[c]{@{}c@{}}QU\\ and\\ AN\end{tabular}}} & \multicolumn{1}{c|}{\textbf{\begin{tabular}[c]{@{}c@{}}QU\\ and\\ DI\end{tabular}}} & \multicolumn{1}{c|}{\textbf{\begin{tabular}[c]{@{}c@{}}QU\\ and\\ JO\end{tabular}}} & \multicolumn{1}{c|}{\textbf{\begin{tabular}[c]{@{}c@{}}QU\\ and\\ SA\end{tabular}}} & \multicolumn{1}{c|}{\textbf{\begin{tabular}[c]{@{}c@{}}QU\\ and\\ SU\end{tabular}}} & \textbf{Total} \\ 
\hline
\textbf{Muslim}      & \multicolumn{1}{c|}{236}                                                              & \multicolumn{1}{c|}{269}                                                              & \multicolumn{1}{c|}{248}                                                              & \multicolumn{1}{c|}{149}                                                              & \multicolumn{1}{c|}{19}                                                               & \multicolumn{1}{c|}{221}                                                              & \multicolumn{1}{c|}{420}                                                              & \multicolumn{1}{c|}{232}                                                              & \multicolumn{1}{c|}{33}                                                               & \multicolumn{1}{c|}{9}                                                                & \multicolumn{1}{c|}{187}                                                              & \multicolumn{1}{c|}{55}                                                               & \multicolumn{1}{c|}{622}                                                              & \multicolumn{1}{c|}{54}                                                               & \multicolumn{1}{c|}{1}                                                                & \multicolumn{1}{c|}{397}                                                              & \multicolumn{1}{c|}{147}                                                              & \multicolumn{1}{c|}{24}                                                               & \multicolumn{1}{c|}{9}                                                                & \multicolumn{1}{c|}{337}                                                              & 3669           \\
\textbf{Women}        & \multicolumn{1}{c|}{108}                                                              & \multicolumn{1}{c|}{159}                                                              & \multicolumn{1}{c|}{161}                                                              & \multicolumn{1}{c|}{71}                                                               & \multicolumn{1}{c|}{10}                                                               & \multicolumn{1}{c|}{151}                                                              & \multicolumn{1}{c|}{181}                                                              & \multicolumn{1}{c|}{152}                                                              & \multicolumn{1}{c|}{14}                                                               & \multicolumn{1}{c|}{10}                                                               & \multicolumn{1}{c|}{59}                                                               & \multicolumn{1}{c|}{44}                                                               & \multicolumn{1}{c|}{378}                                                              & \multicolumn{1}{c|}{26}                                                               & \multicolumn{1}{c|}{1}                                                                & \multicolumn{1}{c|}{258}                                                              & \multicolumn{1}{c|}{91}                                                               & \multicolumn{1}{c|}{15}                                                               & \multicolumn{1}{c|}{11}                                                               & \multicolumn{1}{c|}{134}                                                              & 2034           \\
\textbf{LGBT+}       & \multicolumn{1}{c|}{93}                                                               & \multicolumn{1}{c|}{171}                                                              & \multicolumn{1}{c|}{129}                                                              & \multicolumn{1}{c|}{48}                                                               & \multicolumn{1}{c|}{8}                                                                & \multicolumn{1}{c|}{140}                                                              & \multicolumn{1}{c|}{208}                                                              & \multicolumn{1}{c|}{85}                                                               & \multicolumn{1}{c|}{12}                                                               & \multicolumn{1}{c|}{4}                                                                & \multicolumn{1}{c|}{45}                                                               & \multicolumn{1}{c|}{36}                                                               & \multicolumn{1}{c|}{335}                                                              & \multicolumn{1}{c|}{31}                                                               & \multicolumn{1}{c|}{2}                                                                & \multicolumn{1}{c|}{190}                                                              & \multicolumn{1}{c|}{96}                                                               & \multicolumn{1}{c|}{11}                                                               & \multicolumn{1}{c|}{15}                                                               & \multicolumn{1}{c|}{137}                                                              & 1796           \\
\textbf{Jews}         & \multicolumn{1}{c|}{123}                                                              & \multicolumn{1}{c|}{113}                                                              & \multicolumn{1}{c|}{94}                                                               & \multicolumn{1}{c|}{53}                                                               & \multicolumn{1}{c|}{10}                                                               & \multicolumn{1}{c|}{109}                                                              & \multicolumn{1}{c|}{206}                                                              & \multicolumn{1}{c|}{54}                                                               & \multicolumn{1}{c|}{14}                                                               & \multicolumn{1}{c|}{9}                                                                & \multicolumn{1}{c|}{73}                                                               & \multicolumn{1}{c|}{32}                                                               & \multicolumn{1}{c|}{251}                                                              & \multicolumn{1}{c|}{36}                                                               & \multicolumn{1}{c|}{1}                                                                & \multicolumn{1}{c|}{151}                                                              & \multicolumn{1}{c|}{67}                                                               & \multicolumn{1}{c|}{5}                                                                & \multicolumn{1}{c|}{5}                                                                & \multicolumn{1}{c|}{164}                                                              & 1570           \\
\textbf{Refugee}     & \multicolumn{1}{c|}{10}                                                               & \multicolumn{1}{c|}{8}                                                                & \multicolumn{1}{c|}{37}                                                               & \multicolumn{1}{c|}{12}                                                               & \multicolumn{1}{c|}{3}                                                                & \multicolumn{1}{c|}{10}                                                               & \multicolumn{1}{c|}{22}                                                               & \multicolumn{1}{c|}{33}                                                               & \multicolumn{1}{c|}{5}                                                                & \multicolumn{1}{c|}{0}                                                                & \multicolumn{1}{c|}{5}                                                                & \multicolumn{1}{c|}{4}                                                                & \multicolumn{1}{c|}{55}                                                               & \multicolumn{1}{c|}{6}                                                                & \multicolumn{1}{c|}{0}                                                                & \multicolumn{1}{c|}{27}                                                               & \multicolumn{1}{c|}{11}                                                               & \multicolumn{1}{c|}{3}                                                                & \multicolumn{1}{c|}{1}                                                                & \multicolumn{1}{c|}{28}                                                               & 280            \\
\textbf{AP} & \multicolumn{1}{c|}{7}                                                                & \multicolumn{1}{c|}{10}                                                               & \multicolumn{1}{c|}{6}                                                                & \multicolumn{1}{c|}{5}                                                                & \multicolumn{1}{c|}{1}                                                                & \multicolumn{1}{c|}{9}                                                                & \multicolumn{1}{c|}{16}                                                               & \multicolumn{1}{c|}{4}                                                                & \multicolumn{1}{c|}{0}                                                                & \multicolumn{1}{c|}{0}                                                                & \multicolumn{1}{c|}{4}                                                                & \multicolumn{1}{c|}{1}                                                                & \multicolumn{1}{c|}{22}                                                               & \multicolumn{1}{c|}{2}                                                                & \multicolumn{1}{c|}{0}                                                                & \multicolumn{1}{c|}{11}                                                               & \multicolumn{1}{c|}{4}                                                                & \multicolumn{1}{c|}{1}                                                                & \multicolumn{1}{c|}{2}                                                                & \multicolumn{1}{c|}{11}                                                               & 116            \\
\textbf{IMGT}   & \multicolumn{1}{c|}{99}                                                               & \multicolumn{1}{c|}{101}                                                              & \multicolumn{1}{c|}{260}                                                              & \multicolumn{1}{c|}{87}                                                               & \multicolumn{1}{c|}{15}                                                               & \multicolumn{1}{c|}{147}                                                              & \multicolumn{1}{c|}{137}                                                              & \multicolumn{1}{c|}{244}                                                              & \multicolumn{1}{c|}{29}                                                               & \multicolumn{1}{c|}{5}                                                                & \multicolumn{1}{c|}{64}                                                               & \multicolumn{1}{c|}{22}                                                               & \multicolumn{1}{c|}{441}                                                              & \multicolumn{1}{c|}{32}                                                               & \multicolumn{1}{c|}{3}                                                                & \multicolumn{1}{c|}{242}                                                              & \multicolumn{1}{c|}{76}                                                               & \multicolumn{1}{c|}{30}                                                               & \multicolumn{1}{c|}{14}                                                               & \multicolumn{1}{c|}{200}                                                              & 2248           \\
\textbf{Disable}     & \multicolumn{1}{c|}{30}                                                               & \multicolumn{1}{c|}{51}                                                               & \multicolumn{1}{c|}{65}                                                               & \multicolumn{1}{c|}{28}                                                               & \multicolumn{1}{c|}{1}                                                                & \multicolumn{1}{c|}{33}                                                               & \multicolumn{1}{c|}{75}                                                               & \multicolumn{1}{c|}{59}                                                               & \multicolumn{1}{c|}{7}                                                                & \multicolumn{1}{c|}{1}                                                                & \multicolumn{1}{c|}{12}                                                               & \multicolumn{1}{c|}{17}                                                               & \multicolumn{1}{c|}{125}                                                              & \multicolumn{1}{c|}{18}                                                               & \multicolumn{1}{c|}{1}                                                                & \multicolumn{1}{c|}{100}                                                              & \multicolumn{1}{c|}{18}                                                               & \multicolumn{1}{c|}{6}                                                                & \multicolumn{1}{c|}{7}                                                                & \multicolumn{1}{c|}{42}                                                               & 696            \\
\textbf{PoC}          & \multicolumn{1}{c|}{62}                                                               & \multicolumn{1}{c|}{122}                                                              & \multicolumn{1}{c|}{68}                                                               & \multicolumn{1}{c|}{47}                                                               & \multicolumn{1}{c|}{3}                                                                & \multicolumn{1}{c|}{80}                                                               & \multicolumn{1}{c|}{148}                                                              & \multicolumn{1}{c|}{54}                                                               & \multicolumn{1}{c|}{13}                                                               & \multicolumn{1}{c|}{7}                                                                & \multicolumn{1}{c|}{45}                                                               & \multicolumn{1}{c|}{13}                                                               & \multicolumn{1}{c|}{223}                                                              & \multicolumn{1}{c|}{20}                                                               & \multicolumn{1}{c|}{1}                                                                & \multicolumn{1}{c|}{122}                                                              & \multicolumn{1}{c|}{67}                                                               & \multicolumn{1}{c|}{8}                                                                & \multicolumn{1}{c|}{5}                                                                & \multicolumn{1}{c|}{100}                                                              & 1208           \\
\textbf{Others}       & \multicolumn{1}{c|}{25}                                                               & \multicolumn{1}{c|}{23}                                                               & \multicolumn{1}{c|}{28}                                                               & \multicolumn{1}{c|}{11}                                                               & \multicolumn{1}{c|}{2}                                                                & \multicolumn{1}{c|}{28}                                                               & \multicolumn{1}{c|}{37}                                                               & \multicolumn{1}{c|}{16}                                                               & \multicolumn{1}{c|}{6}                                                                & \multicolumn{1}{c|}{2}                                                                & \multicolumn{1}{c|}{19}                                                               & \multicolumn{1}{c|}{7}                                                                & \multicolumn{1}{c|}{57}                                                               & \multicolumn{1}{c|}{5}                                                                & \multicolumn{1}{c|}{1}                                                                & \multicolumn{1}{c|}{38}                                                               & \multicolumn{1}{c|}{19}                                                               & \multicolumn{1}{c|}{3}                                                                & \multicolumn{1}{c|}{1}                                                                & \multicolumn{1}{c|}{28}                                                               & 356            \\ \hline
\textbf{Total}        & \multicolumn{1}{c|}{793}                                                              & \multicolumn{1}{c|}{1027}                                                             & \multicolumn{1}{c|}{1096}                                                             & \multicolumn{1}{c|}{511}                                                              & \multicolumn{1}{c|}{72}                                                               & \multicolumn{1}{c|}{928}                                                              & \multicolumn{1}{c|}{1450}                                                             & \multicolumn{1}{c|}{933}                                                              & \multicolumn{1}{c|}{133}                                                              & \multicolumn{1}{c|}{47}                                                               & \multicolumn{1}{c|}{513}                                                              & \multicolumn{1}{c|}{231}                                                              & \multicolumn{1}{c|}{2509}                                                             & \multicolumn{1}{c|}{230}                                                              & \multicolumn{1}{c|}{11}                                                               & \multicolumn{1}{c|}{1536}                                                             & \multicolumn{1}{c|}{596}                                                              & \multicolumn{1}{c|}{106}                                                              & \multicolumn{1}{c|}{70}                                                               & \multicolumn{1}{c|}{1181}                                                             & 13973          \\ \hline
\textbf{Train}        & \multicolumn{1}{c|}{505}                                                              & \multicolumn{1}{c|}{656}                                                              & \multicolumn{1}{c|}{686}                                                              & \multicolumn{1}{c|}{325}                                                              & \multicolumn{1}{c|}{39}                                                               & \multicolumn{1}{c|}{605}                                                              & \multicolumn{1}{c|}{912}                                                              & \multicolumn{1}{c|}{544}                                                              & \multicolumn{1}{c|}{84}                                                               & \multicolumn{1}{c|}{30}                                                               & \multicolumn{1}{c|}{294}                                                              & \multicolumn{1}{c|}{142}                                                              & \multicolumn{1}{c|}{1594}                                                             & \multicolumn{1}{c|}{132}                                                              & \multicolumn{1}{c|}{7}                                                                & \multicolumn{1}{c|}{978}                                                              & \multicolumn{1}{c|}{381}                                                              & \multicolumn{1}{c|}{68}                                                               & \multicolumn{1}{c|}{47}                                                               & \multicolumn{1}{c|}{772}                                                              & 8801           \\
\textbf{Dev}          & \multicolumn{1}{c|}{122}                                                              & \multicolumn{1}{c|}{158}                                                              & \multicolumn{1}{c|}{161}                                                              & \multicolumn{1}{c|}{90}                                                               & \multicolumn{1}{c|}{10}                                                               & \multicolumn{1}{c|}{152}                                                              & \multicolumn{1}{c|}{247}                                                              & \multicolumn{1}{c|}{140}                                                              & \multicolumn{1}{c|}{29}                                                               & \multicolumn{1}{c|}{6}                                                                & \multicolumn{1}{c|}{75}                                                               & \multicolumn{1}{c|}{32}                                                               & \multicolumn{1}{c|}{425}                                                              & \multicolumn{1}{c|}{49}                                                               & \multicolumn{1}{c|}{3}                                                                & \multicolumn{1}{c|}{235}                                                              & \multicolumn{1}{c|}{87}                                                               & \multicolumn{1}{c|}{11}                                                               & \multicolumn{1}{c|}{10}                                                               & \multicolumn{1}{c|}{159}                                                              & 2201           \\
\textbf{Test}         & \multicolumn{1}{c|}{166}                                                              & \multicolumn{1}{c|}{213}                                                              & \multicolumn{1}{c|}{249}                                                              & \multicolumn{1}{c|}{96}                                                               & \multicolumn{1}{c|}{23}                                                               & \multicolumn{1}{c|}{171}                                                              & \multicolumn{1}{c|}{291}                                                              & \multicolumn{1}{c|}{249}                                                              & \multicolumn{1}{c|}{20}                                                               & \multicolumn{1}{c|}{11}                                                               & \multicolumn{1}{c|}{144}                                                              & \multicolumn{1}{c|}{57}                                                               & \multicolumn{1}{c|}{490}                                                              & \multicolumn{1}{c|}{49}                                                               & \multicolumn{1}{c|}{1}                                                                & \multicolumn{1}{c|}{323}                                                              & \multicolumn{1}{c|}{128}                                                              & \multicolumn{1}{c|}{27}                                                               & \multicolumn{1}{c|}{13}                                                               & \multicolumn{1}{c|}{250}                                                              & 2971           \\ \hline
\end{tabular}%
}

\caption{Counterspeech distribution in \texttt{MultiCONAN} across various multi-attribute combinations, categorized by target groups (see abbreviations in Section \ref{sec:data}).}
\label{tab:table1_stats}
\vspace{-5mm}
\end{table*}

\section{The \texttt{MultiCONAN} Dataset}\label{sec:data} 
Due to the superior quality of counterspeech (see Appendix \ref{adav_intconan}), we selected \texttt{IntentCONANv2} \cite{hengle2024intentconditioned} as the foundation for our work. \texttt{IntentCONANv2}, while valuable, has limitations that restrict its ability to fully capture the diversity and complexity of counterspeech. Its dependence on a single attribute, `strategy,' oversimplifies responses and fails to account for the emotional tone, which plays a critical role in shaping effective counterspeech. For instance, an informative counterspeech can convey vastly different emotional tones, such as joy or sadness, which are not captured in the original dataset. Building upon this, we introduce \texttt{MultiCONAN}, an enhanced version featuring additional emotion class labels for each counterspeech instance. \texttt{MultiCONAN} includes $13,973$ CS instances of \texttt{IntentCONANv2}, each tagged with one of five emotion classes: Anger (AN), Joy (JO), Disgust (DI), Sad (SA), and Surprise (SU) (See Table \ref{tab:table1_stats}). The added emotion annotations enhance analysis granularity and facilitate the development of models that integrate both strategy and emotional context. This annotation framework enables exploration of how emotional tone interacts with strategic strategy and supports the creation of nuanced CS generation models \cite{gupta-etal-2023-counterspeeches, hengle2024intentconditioned}. \texttt{MultiCONAN} thus serves as a valuable resource for advancing research in CS generation, aiming to produce contextually appropriate and emotionally resonant responses. For detailed information on the annotation process and the statistics of the data set, refer to Appendices \ref{sec:data_annotation},\ref{procedure_annotatioin}, and  \ref{data:statistic}, respectively.

\vspace{-2mm}

\begin{figure*}[t]
    \centering
    \includegraphics[width=0.8\textwidth]{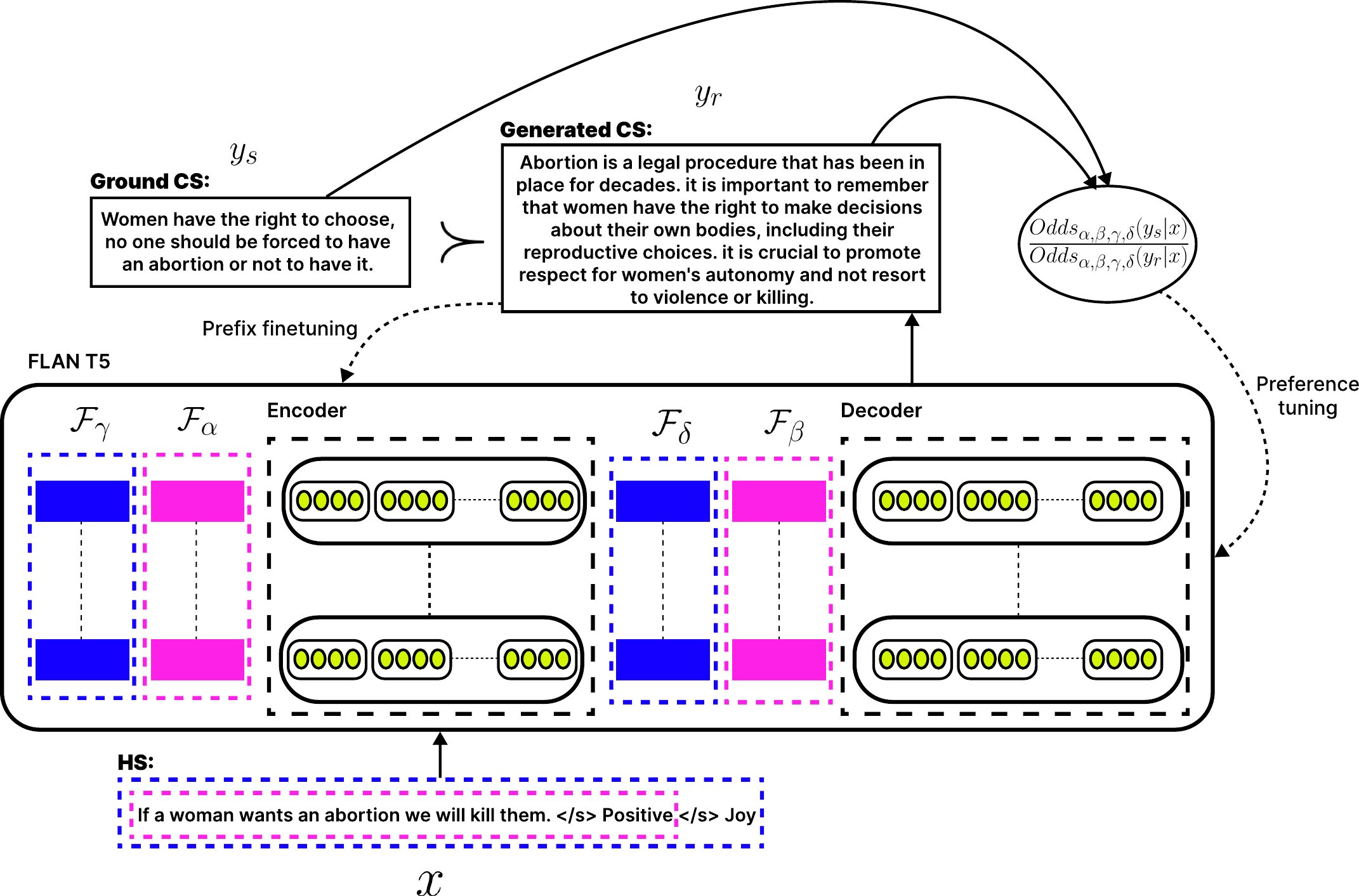}
\caption{Our proposed model, \texttt{HiPPrO}, follows a two-phase pipeline. In Phase 1, we train the prefix parameters ($\alpha, \beta, \gamma, \delta$) associated with prefix adapters ($\mathcal{F_{\alpha}},\mathcal{F_{\beta}}$) on encoder side and ($\mathcal{F_{\gamma}}, \mathcal{F_{\delta}}$) on the decoder side. For $\alpha$ and $\beta$, the model is trained with hate speech and strategy, separated by </s> as input. While training $\gamma$ and $\delta$, the optimal parameters, $\alpha^{*}$ and $\beta^{*}$, are kept fixed, and it includes hate speech, strategy, and emotion, separated by </s> as input $x$, this is also used as a prompt input during the second phase. In Phase 2, we apply preference tuning using an odd ratio loss, where the ground-truth counterspeech serves as the chosen candidate $y_{s}$, and the model-generated counterspeech is treated as the rejected candidate $y_{r}$.} \label{fig:figure1_9}
  \vspace{-5mm}
\end{figure*}
\section{Proposed Methodology}
\label{sec:proposed_methodology}
In this section, we elaborate on the inner workings and structural components of \texttt{\texttt{HiPPrO}}, a novel automated counterspeech generation framework. Here, we explain how our designed model can address the previous challenges by (i) generating multi-attribute conditioned counterspeech that can address hateful comments through \textit{semantic relevancy}, and (ii) aligning it with qualitative human-generated responses through a reward and reference-model-free approach to ensure that the generation is \textit{contrastive} (Figure \ref{fig:figure1_9}). 

\subsection*{Task Formulation}
We use our curated \texttt{MultiCONAN} dataset for generating multi-attribute counterspeech generation. Considering our dataset as  \( \mathcal{D} = \{ (h_1, i_1, e_1, c_1), \ldots, (h_n, i_n, e_n, c_n) \} \), where \( h_i \in \mathcal{H} \) is the \( i \)-th hate speech statement, \( c_i \in \mathcal{C} \) is the counterspeech corresponding to \( h_i \), and \( i_i \in \mathcal{I} \) and \( e_i \in \mathcal{E} \) are the strategy and emotion categories of \( c_i \), respectively. Our objective is to learn a stochastic counterspeech generation function \( \psi : \mathcal{H} \times \mathcal{I} \times \mathcal{E}  \rightarrow \mathcal{C} \), such that \( c_i \sim \psi(\cdot|h_i, i_i,e_i) \).

We address this problem by decomposing the counterspeech generation task into two phases. In the first phase, we focus on learning the prefix vectors for individual attributes and capturing their conditional dependencies using a two-step method. Initially, we learn the prefix vector for the strategy category and collect the optimal prefix vectors. Subsequently, we add another prefix vector initialized with the previously learned values and optimized it for both the strategy and emotion categories while keeping the model parameters from step one fixed. In the second phase, we employ a reward and reference model-free alignment method called the odds ratio preference optimization algorithm \cite{hong2024reference}. During this, we treat the counterspeech generated in phase one as the rejected data column and the actual ground truth counterspeech as the selected data column. For all our experiments, we utilize FLAN-T5 \cite{chung2022scaling} as the base model due to its robust reasoning and multi-task learning capabilities.

\subsection*{Phase 1: Hierarchical Prefix Optimization (HIPO)}
\label{sec:phase1} 

In this section, we mathematically formalize the concept of hierarchical prefix learning. The prefix vectors, represented by tunable key-value pairs \cite{vaswani2017attention}, are introduced in two sub-phases. In the first sub-phase, we add $|V_{\mathcal{I}}|$ virtual prefixes with dimension $d$ across $l$ layers. The \text{prefix adapters}, $\mathcal{F_{\alpha}}$ and $\mathcal{F_{\beta}}$, introduce task-specific continuous vectors, $\alpha, \beta \in \mathbb{R}^{|V_\mathcal{I}| \times l \times 2d}$ to the encoder and decoder, respectively, guiding counterspeech generation according to user strategy.

We maximize the expected log-likelihood and collect the optimal prefix adapters $\mathcal{F_{\alpha^{*}}}, \mathcal{F_{\beta^{*}}}$ for the strategy-guided counterspeech generation, where the hate speech and strategy are sampled from $\mathcal{D}$, as denoted by,
\begin{equation}\small
\begin{aligned}
\mathcal{F_{\alpha^{*}}}, \mathcal{F_{\beta^{*}}} = \underset{\alpha, \beta}{\mathrm{argmax}}\mathop{{}\mathbb{E}_{(h,i\sim \mathcal{D})}} \log (\mathcal{F_{\alpha}}[\pi^{ENC}_{\theta}(X(h,i);\\ \theta,\alpha)],\mathcal{F_{\beta}}[\pi^{DEC}_{\theta}(X(c);\theta,\beta)])
\end{aligned}
\end{equation}

where $\pi^{ENC}_{\theta}$ and $\pi^{DEC}_{\theta}$ are the encoder and decoder part of the model, respectively. The input $X=[h; i; c]$, where $(h,i,c) \in \mathcal{D}$. We further add another set of adapters, $\mathcal{F_{\gamma}}$, and $\mathcal{F_{\delta}}$, on the encoder and decoder side, which add trainable prefix vectors $\gamma, \delta \in \mathbb{R}^{|V_\mathcal{E}| \times l \times 2d}$ on top of the previously trained $\mathcal{F_{\alpha^{*}}}, \mathcal{F_{\beta^{*}}}$, where $|V_{\mathcal{E}}|$ is the number of virtual tokens for second sub-phase. These prefix vectors are responsible for guiding the counterspeech generation with strategy and emotion-specific attributes. Here, the reformulated input is $X^{'}=[h; i; e; c]$, where $(h,i,e,c) \in \mathcal{D}$. During training, we freeze the model parameters $\theta$ and the strategy-specific prefix parameters $\alpha, \beta$ and maximize the expected log-likelihood to get optimal prefix adapters $\mathcal{F_{\gamma^{*}}}, \mathcal{F_{\delta^{*}}}$ for optimal $\gamma^{*}, \delta^{*}$ given by,
\vspace{-2mm}
\begin{equation}\small
\begin{aligned}
\mathcal{F_{\gamma^{*}}}, \mathcal{F_{\delta^{*}}} = \underset{\gamma, \delta}{\mathrm{argmax}}\mathop{{}\mathbb{E}_{(h,i,e\sim \mathcal{D})}} \log (\mathcal{F_{\gamma}}[\mathcal{F_{\alpha^{*}}}   [\pi^{ENC}_{\theta} ( \\ X^{'}(h,i,e); \theta,\alpha^{*},\gamma)]], \mathcal{F_{\delta}}[\mathcal{F_{\beta^{*}}}[\pi^{DEC}_{\theta}(X^{'}(c);\\ \theta,\beta^{*},\delta)]])
\end{aligned}
\end{equation}

Here, all $\alpha^{*}, \beta^{*}, \gamma^{*}$ and $\delta^{*}$ are responsible for the strategy and emotion-conditioned counterspeech generation. Now, our model is ready to advance to Phase 2, where we apply a preference-tuning approach to all pre-trained prefix vectors. This step further optimizes the model to align counterspeech responses more closely with human-generated outputs.

\subsection*{Phase 2: Preference Optimization}
\label{sec:phase2} 
In this phase, we focus on optimizing our model to produce counterspeech instances that are both effective and non-toxic by employing a reward and reference-free preference tuning approach. Inspired by \citet{hong2024reference}, we use the odd ratio method to align the output with the ground truth preferences. This method self-penalizes the model output probabilities using its corresponding odds, denoted as,
\vspace{-5mm}
\begin{equation}
\begin{aligned}
Odds_{\alpha,\beta, \gamma, \delta}(y|x) = \frac{\pi_{\theta, \alpha,\beta, \gamma, \delta}(y|x)}{1-\pi_{\theta, \alpha,\beta, \gamma, \delta}(y|x)}
\end{aligned}
\end{equation}
Now this ratio has an interesting characteristic -- it boils down to less than one when the desired probability is less than its odds, which eventually penalizes the model output and gives a smaller number. The odd ratio is the ratio between two odds of two independent events. The odd ration $OR$ is given by,
\vspace{-3mm}
\begin{equation}
\begin{aligned}
OR_{\alpha,\beta, \gamma, \delta}(y_{s},y_{r}) = \frac{Odds_{\alpha,\beta, \gamma, \delta}(y_{s}|x)}{Odds_{\alpha,\beta, \gamma, \delta}(y_{r}|x)}
\end{aligned}
\end{equation}
Here, $y_{s}$ denotes the ground-truth counterspeech, which is human-generated, and $y_{r}$ denotes the model-generated counterspeech, which is trained during the first stage. $OR_{\alpha,\beta, \gamma, \delta}(y_{s},y_{r})$ becomes high when the numerator is higher than the denominator, and hence $\pi_{\theta,\alpha,\beta, \gamma, \delta}(y_{s}|x)$. 

Let us consider the preference tuning dataset \( \mathcal{D^{'}} = \{ (h_1, y_{s1}, y_{r1}), \ldots, (h_n, y_{sn}, y_{rn}) \} \), where $y_{si}$ and $y_{ri}$ are the $i^{th}$ ground-truth response and model-generated response for a hate speech $h_{i}$. The final loss is the expectation of prefix-tuned loss and odds ratio loss over the data samples sampled from \( \mathcal{D^{'}}\). As shown in Equation \ref{a}, the final loss for preference optimization $J_{final}$ is the combined loss for finetuned loss ($J_{finetuned}$) and odds ratio loss $J_{OR}$ weighted by a factor $\epsilon $.
\vspace{-1mm}
\begin{equation}
\begin{aligned}
J_{final} = \mathop{{}\mathbb{E}_{(x,y_{s},y_{r}\sim \mathcal{D^{'}})}}  [J_{finetuned} + \epsilon J_{OR}]
\end{aligned}
\label{a}
\end{equation}

The $J_{finetuned}$ is simply the negative log-likelihood loss obtained in the prefix-tuning step, and $J_{OR}$ is given by,
\vspace{-3mm}
\begin{equation}
\begin{aligned}
J_{OR} = -\log(\sigma(\log(\frac{Odds_{\alpha,\beta, \gamma, \delta}(y_{s}|x)}{Odds_{\alpha,\beta, \gamma, \delta}(y_{r}|x)})))
\end{aligned}
\label{b}
\end{equation}
where $\sigma$ denotes the sigmoid function. The final objective is to minimize $J_{final}$ by updating the continuous prefix vectors so that the model outputs can align more with the ground-truth counterspeech. For detailed information regarding computational resources and hyperparameter settings, please refer to Appendix \ref{Computing Information} and Appendix \ref{Hyper-parameter Information}, respectively. \footnote{
The model was trained for up to 50 epochs using a callback, with a fixed training batch size and a learning rate of 4 and $1 \times e^{-4}$, respectively.}
\vspace{-2mm}
\begin{table*}[!ht]
\begin{center}
\scalebox{1}
{
\setlength{\tabcolsep}{1mm} 
\fontsize{7pt}{10pt}\selectfont 
\begin{tabular}{lccccccccccc} 
\toprule
\textbf{Method} & \textbf{Prompt/Adapter} & \multicolumn{3}{c}{\textbf{ROUGE} (↑)} & \textbf{M} (↑) & \textbf{BS} (↑) & \textbf{CoSim} (↑) & \textbf{SC} (↑) & \textbf{EC} (↑) & \textbf{TC} (↑) & \textbf{T} (↓) \\ 
\cmidrule(lr){3-5}
~ & ~ & \textbf{R1} & \textbf{R2} & \textbf{RL} & ~ & ~ & ~ & ~ & ~ & ~ & ~ \\ 
\midrule
GPS & \( - \) & \( 0.089 \) & \( 0.011 \) & \( 0.075 \) & \( 0.055 \) & \( 0.840 \) & \( 0.287 \) & \( 0.247 \) & \( 0.292 \) & \( 0.394 \) & \( 0.146 \) \\ 
DialoGPT & \( - \) & \( 0.164 \) & \( 0.050 \) & \( 0.108 \) & \( 0.198 \) & \( 0.749 \) & \( 0.536 \) & \underline{\( 0.714 \)} & \( 0.571 \) & \( 0.969 \) & \( 0.309 \) \\ 
CoARL & \( - \) & \( 0.156 \) & \( 0.035 \) & \( 0.126 \) & \( 0.110 \) & \( 0.860 \) & \( 0.440 \) & \( 0.380 \) & \( 0.468 \) & \( 0.938 \) & \( 0.258 \) \\ 
\midrule
Vanilla FLAN-T5$_{\mathrm{XXL}}$ & ZS & \( 0.176 \) & \( 0.045 \) & \( 0.145 \) & \( 0.118 \) & \( 0.866 \) & \( 0.479 \) & \( 0.391 \) & \( 0.586 \) & \( \mathbf{0.989} \) & \( 0.341 \) \\ 
Vanilla FLAN-T5$_{\mathrm{XXL}}$ & FS & \( 0.174 \) & \( 0.042 \) & \( 0.142 \) & \( 0.116 \) & \( 0.864 \) & \( 0.456 \) & \( 0.352 \) & \( 0.478 \) & \( 0.976 \) & \( 0.301 \) \\ 

\midrule
GPT-3.5-Turbo & ZS & \( 0.239 \) & \( 0.052 \) & \( 0.160 \) & \( \mathbf{0.249} \) & \( 0.868 \) & \( \mathbf{0.585} \) & \( 0.533 \) & \underline{\( 0.783 \)} & \underline{\( 0.978 \)} & \( \mathbf{0.0007} \) \\ 
GPT-3.5-Turbo & FS & \underline{\( 0.242 \)} & \underline{\( 0.059 \)} & \underline{\( 0.161 \)} & \( \mathbf{0.249} \) & \( 0.868 \) & \( 0.564 \) & \( 0.309 \) & \( 0.527 \) & \( 0.888 \) & \underline{\( 0.012 \)} \\ 
\midrule
GPT-4 & ZS & \( 0.221 \) & \( 0.037 \) & \( 0.145 \) & \( 0.229 \) & \( 0.864 \) & \underline{\( 0.583 \)} & \( 0.545 \) & \( \mathbf{0.790} \) & \( 0.976 \) & \( 0.030 \) \\ 
GPT-4 & FS & \( 0.226 \) & \( 0.039 \) & \( 0.149 \) & \( 0.213 \) & \( 0.868 \) & \( 0.554 \) & \( 0.454 \) & \( 0.515 \) & \( 0.932 \) & \( 0.014 \)  \\ 
\midrule
FLAN-T5$_{\mathrm{XXL}}$ & Retrieval-based & \(0.188\) & \(0.031\) & \(0.132\) & \(0.160\) & \(0.862\) & \(0.494\) & \(0.379\) & \(0.549\) & \(0.907\) & \(0.097\) \\
GPT2$_{\mathrm{XL}}$ & Retrieval-based & \(0.113\) & \(0.011\) & \(0.074\) & \(0.154\) & \(0.821\) & \(0.366\) & \(0.309\) & \(0.441\) & \(0.754\) & \(0.103\) \\
Llama-3.1-8B-Instruct & Retrieval-based & \(0.128\) & \(0.026\) & \(0.086\) & \(0.197\) & \(0.828\) & \(0.462\) & \(0.347\) & \(0.678\) & \(0.891\) & \(0.096\) \\
Mistral-7B-Instruct-v0.2 & Retrieval-based & \(0.166\) & \(0.033\) & \(0.107\) & \(0.224\) & \(0.847\) & \(0.530\) & \(0.356\) & \(0.782\) & \(0.932\) & \(0.068\) \\
DeepSeek-R1-Distill-Llama-8B & Retrieval-based & \(0.132\) & \(0.027\) & \(0.086\) & \(0.210\) & \(0.830\) & \(0.471\) & \(0.373\) & \(0.751\) & \(0.890\) & \(0.059\) \\

\midrule
Vanilla FLAN-T5 $_{\mathrm{XXL}}$ & PrefixTuning & \( 0.229 \) & \( 0.052 \) & \( 0.158 \) & \( 0.222 \) & \underline{\( 0.870 \)} & \( 0.539 \) & \( 0.470 \) & \( 0.666 \) & \( 0.901 \) & \( 0.024 \) \\ 
Vanilla BART $_{\mathrm{Large}}$ & PrefixTuning & \( 0.207 \) & \( 0.042 \) & \( 0.131 \) & \( 0.226 \) & \( 0.861 \) & \( 0.226 \) & \( 0.249 \) & \( 0.453 \) & \( 0.915 \) & \( 0.030 \) \\ 
Vanilla GPT2 $_{\mathrm{XL}}$ & PrefixTuning & \( 0.155 \) & \( 0.029 \) & \( 0.122 \) & \( 0.084 \) & \( 0.820 \) & \( 0.487 \) & \( 0.453 \) & \( 0.455 \) & \( 0.937 \) & \( 0.837 \) \\
Vanilla Llama 3.1 Instruct $_{\mathrm{8B}}$ & PrefixTuning & \( 0.168 \) & \( 0.043 \) & \( 0.135 \) & \( 0.093 \) & \( 0.854 \) & \( 0.515 \) & \( 0.552 \) & \( 0.592 \) & \( 0.903 \) & \( 0.605 \) \\
Vanilla Mistral Instruct $_{\mathrm{7B}}$ & PrefixTuning & \( 0.170 \) & \( 0.046 \) & \( 0.140 \) & \( 0.117 \) & \( 0.858 \) & \( 0.510 \) & \( 0.516 \) & \( 0.597 \) & \( 0.920 \) & \( 0.645 \) \\
DeepSeek-R1-Distill-Llama-8B & PrefixTuning & \( 0.035 \) & \( 0.0002 \) & \( 0.034 \) & \( 0.010 \) & \( 0.808 \) & \( 0.181 \) & \( 0.261 \) & \( 0.289 \) & \( 0.582 \) & \( 0.023 \) \\
DeepSeek-llm-7b-chat & PrefixTuning & \( 0.040 \) & \(\ 0.001 \) & \( 0.035 \) & \( 0.027 \) & \( 0.805 \) & \( 0.144 \) & \( 0.261 \) & \( 0.311 \) & \( 0.484 \) & \( 0.019 \) \\
\midrule
\textbf{\texttt{HiPPrO} $_{\mathrm{\textbf{VT=3}}}$ (Ours)} & PrefixTuning & \( \mathbf{0.273^{*}} \) & \( \mathbf{0.082^{*}} \) & \( \mathbf{0.199^{*}} \) & \underline{\( 0.242 \)} & \( \mathbf{0.879^{*}}\) & \( 0.567 \) & \( \mathbf{0.929^{*}} \) & \( 0.706 \) & \( 0.897 \) & \( 0.087 \) \\ 
\midrule
- With out ORPO & PrifixTuning & \( 0.272 \) & \( 0.081 \) & \( 0.198 \) & \( 0.241 \) & \( 0.879 \) & \( 0.567 \) & \( 0.928 \) & \( 0.705 \) & \( 0.896 \) & \( 0.111 \) \\ 
- With DPO & PrefixTuning & \( 0.272 \) & \( 0.081 \) & \( 0.198 \) & \( 0.240 \) & \( 0.879 \) & \( 0.566 \) & \( 0.928 \) & \( 0.685 \) & \( 0.895 \) & \( 0.089 \) \\ 
- HIPO $_{\mathrm{VT=5}}$ & PrifixTuning & \( 0.275 \) & \( 0.081 \) & \( 0.200 \) & \( 0.241 \) & \( 0.880 \) & \( 0.578 \) & \( 0.921 \) & \( 0.656 \) & \( 0.916 \) & \( 0.096 \) \\ 
- HIPO $_{\mathrm{VT=7}}$ & PrefixTuning & \( 0.273 \) & \( 0.084 \) & \( 0.199 \) & \( 0.24 \) & \( 0.879 \) & \( 0.573 \) & \( 0.937 \) & \( 0.643 \) & \( 0.888 \) & \( 0.084 \) \\ 
- HIPO $_{\mathrm{VT=10}}$ & PrefixTuning & \( 0.271 \) & \( 0.081 \) & \( 0.197 \) & \( 0.244 \) & \( 0.878 \) & \( 0.564 \) & \( 0.905 \) & \( 0.630 \) & \( 0.858 \) & \( 0.077 \) \\ 
\midrule
\multicolumn{2}{c}{\( \Delta_{\mathrm{HiPPrO (Ours)} - \mathrm{Best Baseline Method}} \)} & 
\textcolor{ForestGreen}{\( \uparrow0.034 \)} & 
\textcolor{ForestGreen}{\( \uparrow0.023 \)} & 
\textcolor{ForestGreen}{\( \uparrow0.038 \)} & 
\textcolor{red}{\( \downarrow0.007 \)} & 
\textcolor{ForestGreen}{\( \uparrow0.009 \)} & 
\textcolor{red}{\( \downarrow0.018 \)} & 
\textcolor{ForestGreen}{\( \uparrow0.384 \)} & 
\textcolor{red}{\( \downarrow0.084 \)} & 
\textcolor{red}{\( \downarrow0.092 \)} & 
\textcolor{red}{\( \downarrow0.086 \)} \\
\bottomrule
\end{tabular}
}
\end{center}

\caption{Comparing \texttt{HiPPrO} with baselines across various evaluation metrics. Here, ↑ (\textit{resp.} ↓) denotes that higher (\textit{resp.} lower) is better.  \textbf{Bold} (\textit{resp.} \underline{underline}) indicates the best (\textit{resp.} second-ranked) performance. $*$ shows our model \textit{significantly} outperforms (\(p < 0.05\)) the best baselines -- GPT-3.5-Turbo ZS and FS (see Appendix \ref{stest}).}
\label{tab:results_table}
\vspace{-5mm}
\end{table*}

\section{Experimental Setup}
\subsection{Baselines}
\label{sec:baselines}

To evaluate the efficacy of various models, we investigate \textbf{Generate Prune Select (GPS)} \cite{zhu-bhat-2021-generate}, a three-stage pipeline encompassing autoencoding, grammatical filtering, and response selection. Furthermore, we optimize \textbf{DialoGPT} \cite{zhang2020dialogpt} to produce contextually coherent responses. Additionally, we incorporate \textbf{CoARL} \cite{hengle2024intentconditioned}, the state-of-the-art strategy-conditioned CS generation approach. Furthermore, we explore prefix tuning on \textbf{Vanilla FLAN-T5$_{\mathrm{XXL}}$}, and \textbf{BART$_{\mathrm{Large}}$} (excluding \texttt{ HiPPrO}) and conduct experiments with \text{HIPO} (without preference optimization) using different virtual token (VT) sizes (VT = $3, 5, 7, 10$),
VT = $3$ emerging as the optimal configuration in terms of both parameter efficiency ($589,824$ trainable parameters, $0.0052\%$ of total model parameters) and performance metrics. We consider VT = $3$ for preference tuning. With same setup, we experiment with decoder-only models like  \textbf{GPT2$_{\mathrm{XL}}$}, \textbf{Llama 3.1 Instruct$_{\mathrm{8B}}$} \cite{grattafiori2024llama3herdmodels}, \textbf{Mixtral Instruct$_{\mathrm{7B}}$} \cite{jiang2024mixtralexperts}, \textbf{DeepSeek-R1-Distill-Llama-8B} and \textbf{DeepSeek-llm-7b-chat} \cite{deepseekai2025deepseekv3technicalreport}. We also utilize DPO to enhance our evaluation further. Our comprehensive assessment encompasses zero-shot and few-shot performances on three LLMs: \textbf{Vanilla FLAN-T5$_{\mathrm{XXL}}$} \cite{chung2022scaling}, \textbf{GPT-3.5-Turbo} (ChatGPT), and \textbf{GPT-4} \cite{ouyang2022-instuctGPT} (see Appendix \ref{ZS} and \ref{FS}) and simple retrieval-based methods using state-of-the-art open-source LLMs, employing `faiss'\cite{douze2025faisslibrary} as the retrieval method to retrieve the top five training counterspeech examples (see Appendix \ref{RB}).

\subsection{Evaluation Metrics}

Evaluating CS generation presents challenges due to its dynamic nature, diverse response possibilities, and absence of standardized metrics \cite{chung-et-al-2023-understanding-counterspeech}, prompting our framework to utilize comprehensive, multi-dimensional evaluation metrics. These evaluation metrics include lexical similarity, semantic similarity or relevance, strategy conformity, emotion conformity, target conformity, and toxicity score. \textit{Lexical similarity} is evaluated using \textbf{Rouge} \cite{rouge-score-lin-2004} and \textbf{Meteor (M)} \cite{meteor-score-banerjee-lavie-2005}, which quantify the linguistic alignment between generated and reference texts. \textit{Semantic relevance} is assessed with \textbf{cosine similarity (CoSim)} \cite{sentence-transformers} and \textbf{BERTScore (BS)} \cite{bert-score-zhang2020}, ensuring that generated CS engages meaningfully with the primary topic of hate speech. A low relevance score implies a lack of topical coherence, where the CS fails to adequately address the primary subject of hate speech. We also evaluate the effectiveness of incorporating strategic, emotional resonance, and target alignment through \textbf{Strategy Conformity (SC)} \cite{gupta-etal-2023-counterspeeches}, \textbf{Emotion Conformity (EC)}, and \textbf{Target Conformity (TC)}, respectively. These metrics are particularly valuable in scenarios where ground-truth counterspeech is unavailable for unseen hate speech instances. We first train three distinct RoBERTa-large models on our dataset to measure the SC, EC, and TC scores. The models achieve testing accuracy of $0.86$, $0.75$, and $0.88$, respectively, before being considered as evaluation metrics. \textbf{\textit{Toxicity}} (T)\footnote{\href{3}{https://www.perspectiveapi.com/}} levels of generated CS using the \cite{Detoxify} library, ensure that our approach promotes respectful and safe communication.

\section{Experimental Results}

This section presents a comprehensive empirical analysis that systematically evaluates the efficacy of {\texttt{HiPPrO}} in comparison to existing state-of-the-art techniques.

\subsection{Quantitative Results}
Table \ref{tab:results_table} demonstrates the quantitative evaluation across various metrics. \texttt{HiPPrO} shows a notable improvement over baselines across several evaluation metrics. \texttt{HiPPrO} achieves  $0.273$ ROUGE-1, $0.082$ ROUGE-2,  $0.199$ ROUGE-L, substantially higher than GPS, DialoGPT, and CoARL, with an average improvement of $0.117$ in ROUGE-1, $0.05$ in ROUGE-2 and $0.096$ in ROUGE-L. This indicates that
 
\texttt{HiPPrO}'s counterspeech better aligns with reference content in terms of coverage and detail. In BERTScore, \texttt{HiPPrO} scores $0.879$, surpassing GPS ($0.840$), DialoGPT ($0.749$), and CoARL ($0.860$), GPT-4 Few-Shot (FS) ($0.868$) and GPT-3.5 Few-Shot ($0.868$), which indicates {\tt HiPPrO}'s proficiency in preserving the underlying meaning of the original CS. \texttt{HiPPrO} exhibits a slight decrease in CoSim compared to models like GPT-3.5-Turbo Zero-Shot (ZS) with a score of $0.585$, indicating a potential trade-off between contextual relevance and semantic alignment with reference responses.

The generated CS should ideally be a balanced generation of both strategy and emotional attributes, ensuring that responses not only address the harmful content effectively but also align with strategy and emotional tone along with the target audience. However, the evaluation results indicate that it is a challenge for many models to achieve this balance. For instance, while GPT-3.5-Turbo ZS demonstrates a low SC score of $0.533$, its EC score is significantly high at $0.783$, highlighting the difficulty in generating CS that is both strategy-aligned and emotionally resonant. Similarly, GPT-4 ZS shows a high EC score of $0.790$, yet its SC score of $0.545$ suggests that even more advanced LLMs struggle to maintain emotional alignment in the zero-shot and few-shot scenarios. In contrast, \texttt{HiPPrO} demonstrates a better balance, achieving an SC score of $0.929$ and an EC score of $0.706$. \texttt{HiPPrO}'s SC score is substantially higher, indicating its superior ability to generate strategy-aligned CS while still maintaining a better level of emotional resonance. This suggests that \texttt{HiPPrO} is more adept at maintaining the stability between these two critical attributes during the generation process.

The relationship between TC and BS provides insights into how models generate counterspeech; a high TC score with a relatively low BS indicates that the CS is target-specific but may lack deeper semantic alignment with the original CS. For instance, GPT-3.5-Turbo in the ZS setting achieves a TC score of $0.978$ and a BS score of $0.868$, suggesting that while it effectively mentions the target group, it may produce overly generic responses that fail to address nuanced aspects of hate speech. In contrast, \texttt{HiPPrO} manages to strike a better balance by achieving a TC score of $0.897$, which, while slightly lower than GPT-4 few-shot, is complemented by a BS of $0.879$, the highest among all models. This indicates that \texttt{HiPPrO} is not only attentive to the target group but also maintains a strong semantic connection to the original content, making its CS both specific and contextually relevant. Furthermore, when examining the toxicity scores, \texttt{HiPPrO} achieves a respectable score of $0.087$, which, although slightly higher than GPT-3.5-Turbo Zero-Shot, still indicates a significantly low presence of harmful language comparable to all other baselines. This suggests that \texttt{HiPPrO} effectively balances generating contextually rich CS while keeping the content non-toxic and constructive. Statistical analyses reveal that our model significantly outperforms both GPT-3.5 ZS and FS across most metrics, with exceptions in meteor, CoSim, and TC scores. Please refer to Appendix (Section \ref{stest}) for additional information about statistical significance tests.

\subsection{Ablation Study} 

Our ablation study assesses the effects of several components of \texttt{HiPPrO} (Table \ref{tab:results_table}). One such study involves fine-tuning \texttt{HiPPrO} without the ORPO component. The results show that the omission of ORPO leads to a decrease in performance in almost all evaluation metrics, with a notable reduction in the toxicity score by $0.024$. This suggests that ORPO contributes significantly to enhancing both the overall performance and the generation of less toxic counterspeech. Additionally, we fine-tune \texttt{HiPPrO} with DPO instead of ORPO and observe a slight degradation in performance. Although the difference is marginal, the key advantage of ORPO over DPO lies in ORPO's ability to operate without a reference model, which DPO requires. Additionally, we investigate how different numbers of VT in \texttt{HiPPrO} impact its performance, testing configurations with VT= $5, 7,$ and $10$. The results indicate that there is no notable performance gain after VT=$5$. While higher virtual tokens lead to a slight reduction in the toxicity score, the improvements are insufficient across other metrics, suggesting no significant returns with increased token counts.

Our ablation studies also include evaluations of Vanilla FLAN-T5$_{\mathrm{XXL}}$ and Vanilla BART$_{\mathrm{Large}}$, both with prefix tuning, without hierarchical learning, providing further insights into \texttt{HiPPrO}’s effectiveness. We observe that while these models offer competitive performance, they do not surpass \texttt{HiPPrO} in several key metrics. For Vanilla FLAN-T5$_{\mathrm{XXL}}$, prefix tuning yields ROUGE-1, ROUGE-2, and ROUGE-L scores of $0.229$, $0.052$, and $0.158$, respectively. These scores are slightly lower than those of \texttt{HiPPrO}, indicating that while the prefix-tuned FLAN-T5$_{\mathrm{XXL}}$ performs well, \texttt{HiPPrO}’s output is more detailed and comprehensive. However, the SC scores for Vanilla FLAN-T5$_{\mathrm{XXL}}$ and BART$_{\mathrm{Large}}$ are $0.470$ and $0.249$, respectively, while their EC scores are $0.666$ and $0.453$, respectively. \texttt{HiPPrO} achieves higher SC and EC scores ($0.929$ and $0.706$ respectively), highlighting the limitations of simple prefix-tuning methods in effectively aligning CS with multiple attributes.

\subsection{Human Evaluation}
Previous studies \citep{jones-etal-2024-multi,wang-etal-2023-chatgpt,hengle2025csevalautomatedmultidimensionalreferencefree} emphasize the need for a dual evaluation framework, as automatic metrics show weak correlation with human judgments of counterspeech effectiveness. A comprehensive human evaluation was conducted on a random subset of $30$ responses from the top-performing CS generation methods (\texttt{HiPPrO}, CoARL, Few-Shot GPT-4, and GPT-3.5 Turbo) with a random seed value of $1$, ensuring uniform distribution across strategies and emotions. A diverse panel of 35 experts in NLP and social sciences (aged 20-35, 45\% male, 55\% female) evaluated and ranked these responses based on several key metrics. We followed \citet{hengle2024intentconditioned} for our human evaluation. The evaluation framework consists of five key metrics: Independent Counterspeech (ICS) to gauge the response's self-sufficiency; Adequacy (Ad) to assess its linguistic quality; Contextual Relevance (CoRl) to measure its responsiveness to hate speech components; and Argumentative Effectiveness (ArgE) to evaluate its carefulness and convincing. We present the comparative performance of \texttt{HiPPrO} against leading methods through \textit{Win Rate} scores (see Table \ref{tab:human_eval}). For ICS, \texttt{HiPPrO} outperforms the baselines with win rates of $0.91$ and $0.89$ over GPT-4 and GPT-3.5, showing its ability to generate CS that operates effectively without needing extra context. Regarding Ad, \texttt{HiPPrO} achieves higher scores of $0.85$ and $0.87$, reflecting the superior grammatical accuracy and fluency of CS. In terms of CoRl, \texttt{HiPPrO}'s win rates of $0.89$ and $0.87$ highlight its strength in addressing crucial aspects of hate speech, such as targeted biases. Finally, for ArgE, \texttt{HiPPrO} leads with scores of $0.89$ and $0.90$, indicating its effectiveness in delivering compelling and well-structured CS. These results collectively show \texttt{HiPPrO}'s robust performance across all metrics compared to the baseline models.

\begin{table}[t!]
\begin{center}
\resizebox{\columnwidth}{!}{%
\begin{tabular}{l|c|c|c|c} 
\toprule
\multicolumn{1}{l|}{\textbf{Models on comparison}} & \multicolumn{4}{c}{\textbf{Metrics}} \\ 
\midrule
 & \textbf{ICS $\uparrow$} & \textbf{Ad $\uparrow$} & \textbf{CoRl $\uparrow$} & \textbf{ArgE $\uparrow$} \\ 
\midrule
\texttt{HiPPrO} vs CoARL & $0.96$ & $0.98$ & $0.93$ & $0.97$ \\
\texttt{HiPPrO} vs GPT-4 (FS) & $0.91$ & $0.85$ & $0.89$ & $0.89$ \\
\texttt{HiPPrO} vs GPT-3.5 (FS) & $0.89$ & $0.87$ & $0.87$ & $0.90$ \\
\bottomrule
\end{tabular}%
}
\end{center}
\vspace{-2mm}
\caption{Results of the human evaluation study, where responses generated by \texttt{HiPPrO} are shown against those produced by (a) CoARL, (b) GPT-4 (FS), and (c) GPT-3.5 (FS). The results are reported in terms of \text{Win Rate \%}, indicating the \% of instances where \texttt{HiPPrO} outperforms the respective baselines.}
\label{tab:human_eval}
\vspace{-5mm}
\end{table}

\section{Conclusion}

This study presented \texttt{HiPPrO}, a novel two-stage framework for generating controllable, multi-attributed counterspeech. The initial stage comprised a hierarchical learning process, where the model acquired attribute-specific prefixes, thereby guiding the LLM towards targeted counterspeech generation. The subsequent stage involved refining the outputs to enhance their human-like quality and non-toxicity, employing a reward- and reference-free alignment approach. Additionally, we introduced the \texttt{multiCONAN} dataset with strategy- and emotion-specific counterspeech. An extensive evaluation, incorporating a range of quantitative and qualitative measures, demonstrated \texttt{HiPPrO}'s superiority over multiple baselines.

\section*{Limitation}

Our research presents several limitations that warrant consideration. Firstly, the dataset utilized for hate speech and counterspeech is not comprehensive, potentially omitting various forms and targets of online hate. Secondly, the framework's reliance on pre-trained models may introduce inherent biases or inaccuracies stemming from these source models. Additionally, the evaluation metrics employed do not fully align with human perceptions of counterspeech quality, thereby failing to capture the intricate nuances of natural language interactions. Moreover, our framework does not address the possibility of feedback loops or escalation that could arise following the generation of counterspeech, which may influence the long-term effectiveness and impact of our approach. Lastly, while efforts were made to maintain high-quality annotations for counterspeech, it is conceivable that our dataset may not match the caliber of those annotated by more experienced operators from NGOs, such as those found in the Multi-Target CONAN \cite{fanton2021humanintheloop} dataset. Future research could mitigate these limitations by expanding and diversifying the dataset, enhancing the evaluation criteria, and incorporating dialogue modeling into the framework. Our primary focus was on generating effective and non-toxic counterspeech, following prior work \cite{hengle2024intentconditioned}. While we did not explicitly analyze potential biases in this study, we acknowledge that large language models trained on social media data can amplify biases. Addressing and mitigating such biases is indeed a critical area of research.
\section*{Ethics Statement}
We recognize the sensitivity required in addressing online hate speech and acknowledge the ethical and moral complexities inherent in conducting research in this area. This initiative serves as an initial attempt to compile a comprehensive and varied collection of counterspeech responses for each instance of hate speech encountered. We understand that algorithms developed for automated counterspeech may generate responses that fail to accurately convey the intended meanings, highlighting the urgent need to better integrate real-world knowledge into these systems. Despite the potential of generative algorithms, there remains a critical necessity for a robust and diverse database of counterspeech to ensure consistently favorable outcomes. Furthermore, while fully operational counterspeech algorithms have yet to be realized, organizations such as United Against Hate play a crucial role in mitigating the prevalence of hate speech in online environments.

\section{Acknowledgments}
We extend our gratitude to the central HPC facility (Padum) at IIT Delhi for computing. We also sincerely thank Logically and Anusandhan National Research Foundation (DST-NSF Grant: DST/INT/USA/NSF-DST/Tanmoy/P-2/2024) for financial support. Tanmoy acknowledges the support of Rajiv Khemani Young Faculty Chair Professorship in Artificial Intelligence. 

\bibliography{anthology}

\begin{thebibliography}{59}
\bibitem[{Leetaru(2019)}]{leetaru-2019}
Kalev Leetaru. 2019.
\newblock Online toxicity is as old as the web itself but the return to communities may help. forbes magazin.
\end{thebibliography}

\section{Appendix}
\label{sec:appendix-dataset}

\subsection{Annotation Process}
\label{sec:data_annotation}
Our annotation process was conducted by a team of five expert annotators with backgrounds in social science and computational linguistics, specializing in online hate speech and counter-narrative generation. Each annotator had a strong foundation in hate speech analysis, having published research papers or completed advanced studies in the field. To ensure consistency and high-quality annotations, we implemented an extensive training program. This included reviewing existing counter-narrative frameworks, discussing annotation guidelines, and participating in exercises to align understanding of emotion categories. This preparatory phase was crucial for achieving reliable inter-annotator agreement and maintaining the integrity of the annotations.

The annotation process itself followed a rigorous three-phase protocol. Initially, all five annotators independently labeled a common set of $250$ instances to establish a baseline agreement. For evaluating inter-annotator agreement, we utilized both Cohen’s Kappa \cite{cohen1960coefficient} and Fleiss' Kappa \cite{fleiss1971measuring}. Cohen’s Kappa measures pairwise agreement among annotators, finding most values exceeded $0.70$, with several above $0.80$, indicating high consistency (See Table \ref{tab:IAA}). Instances with significant disagreement were addressed through group discussions to align understanding and resolve discrepancies. In the second phase, larger batches were annotated with periodic cross-validation, where $20$\% of instances were randomly assigned to multiple annotators to ensure consistency. This phase involved annotating batches of $2,500, 2,800$, and $3,000$ instances, respectively. Table \ref{Batchinterannotator} shows the batch-wise intra-annotation agreement results. The final phase allowed independent annotation after achieving strong agreement, as measured by Cohen's Kappa exceeding $0.8$. Throughout the process, annotators utilized a custom interface that systematically displayed the existing counterspeech from \texttt{IntentCONANv2}, and emotion category options. To further check the annotation quality, we conducted a post-analysis on the intra-annotator agreement for a randomly selected subset of $5,000$ counterspeech instances (see Table \ref{final_5000}). This analysis was in response to suggestions for additional quality checks. Our study utilized high-quality counterspeech instances from \texttt{IntentCONANv2}, with annotators independently assigning emotion labels based on established guidelines without access to the strategy categories. The resulting uniform distribution of strategy-emotion pairs across target groups was a posterior outcome of this independent process, uninfluenced by pre-existing constraints or biases. These metrics collectively show the robustness and reliability of the \texttt{MultiCONAN} dataset.

\begin{table*}[h!]

\centering

\begin{tabular}{l|c|c|c|c|c}
\toprule
\multicolumn{1}{l|}{\textbf{Batch}} & \multicolumn{5}{c}{\textbf{Inter-annotator Agreement}} \\
\midrule
 & \textbf{Annotator 1} & \textbf{Annotator 2} & \textbf{Annotator 3} & \textbf{Annotator 4} & \textbf{Annotator 5} \\
\midrule
\multicolumn{6}{c}{\textbf{Batch 1 (2500)}} \\
\midrule
Annotator 1 & 1.0 & 0.808783 & 0.811652 & 0.782824 & 0.762829 \\
Annotator 2 & 0.808783 & 1.0 & 0.661458 & 0.63632 & 0.612693 \\
Annotator 3 & 0.811652 & 0.661458 & 1.0 & 0.640035 & 0.619542 \\
Annotator 4 & 0.782824 & 0.63632 & 0.640035 & 1.0 & 0.605464 \\
Annotator 5 & 0.762829 & 0.612693 & 0.619542 & 0.605464 & 1.0 \\
\midrule
\multicolumn{6}{c}{\textbf{Batch 2 (2800)}} \\
\midrule
Annotator 1 & 1.0 & 0.852106 & 0.813067 & 0.817208 & 0.800127 \\
Annotator 2 & 0.852106 & 1.0 & 0.693793 & 0.693689 & 0.687102 \\
Annotator 3 & 0.813067 & 0.693793 & 1.0 & 0.665585 & 0.654362 \\
Annotator 4 & 0.817208 & 0.693689 & 0.665585 & 1.0 & 0.659997 \\
Annotator 5 & 0.800127 & 0.687102 & 0.654362 & 0.659997 & 1.0 \\
\midrule
\multicolumn{6}{c}{\textbf{Batch 3 (3000)}} \\
\midrule
Annotator 1 & 1.0 & 0.942593 & 0.894511 & 0.888579 & 0.85598 \\
Annotator 2 & 0.942593 & 1.0 & 0.840615 & 0.836788 & 0.80862 \\
Annotator 3 & 0.894511 & 0.840615 & 1.0 & 0.796064 & 0.765929 \\
Annotator 4 & 0.888579 & 0.836788 & 0.796064 & 1.0 & 0.758911 \\
Annotator 5 & 0.85598 & 0.80862 & 0.765929 & 0.758911 & 1.0 \\
\bottomrule
\end{tabular}
\caption{Inter-annotator Agreement Coefficients for Each Batch}
\label{Batchinterannotator}
\end{table*}
\begin{table*}[h!]
\centering

\begin{tabular}{l|c|c|c|c|c}
\toprule
\textbf{Annotator} & \textbf{Annotator 1} & \textbf{Annotator 2} & \textbf{Annotator 3} & \textbf{Annotator 4} & \textbf{Annotator 5} \\
\midrule
Annotator 1 & 1.0 & 0.930857 & 0.889566 & 0.855752 & 0.883054 \\
Annotator 2 & 0.930857 & 1.0 & 0.828692 & 0.798059 & 0.823054 \\
Annotator 3 & 0.889566 & 0.828692 & 1.0 & 0.763612 & 0.786144 \\
Annotator 4 & 0.855752 & 0.798059 & 0.763612 & 1.0 & 0.755936 \\
Annotator 5 & 0.883054 & 0.823054 & 0.786144 & 0.755936 & 1.0 \\
\bottomrule
\end{tabular}
\caption{Inter-annotator Agreement of $5000$ Counterspeech Instances for Post Quality Assessment}
\label{final_5000}
\end{table*}
\begin{table}[t!]
    \centering
\setlength{\tabcolsep}{1mm} 
\fontsize{9pt}{12pt}\selectfont
    \begin{tabular}{l|c|c|c|c|c}
        \hline
        & \textbf{A1} & \textbf{A2} & \textbf{A3} & \textbf{A4} & \textbf{A5} \\
        \hline
        \textbf{A1} & $0.000$ & $0.875$ & $0.883$ & $0.863$ & $0.850$ \\
        \textbf{A2} & $0.875$ & $0.000$ & $0.788$ & $0.734$ & $0.748$ \\
        \textbf{A3} & $0.883$ & $0.788$ & $0.000$ & $0.737$ & $0.785$ \\
        \textbf{A4} & $0.808$ & $0.734$ & $0.737$ & $0.000$ & $0.691$ \\
        \textbf{A5} & $0.844$ & $0.748$ & $0.785$ & $0.691$ & $0.000$ \\
        \bottomrule
    \end{tabular}
    \caption{Cohen's Kappa Matrix for Inter-Annotator Agreement among five annotators:  A1, A2, A3, A4, and A5.}
    \label{tab:IAA}
    \vspace{-5mm}
\end{table}

\subsection{Procedure and Annotation Criteria}
\label{procedure_annotatioin}
Before beginning the annotation process, all annotators thoroughly reviewed the field guide on ``addressing online harassment'' \footnote{\url{https://onlineharassmentfieldmanual.pen.org/}}. This preparatory phase involved extensive discussions with the annotators to deepen their understanding of counterspeech. These dialogues ensured that the annotators were well-equipped with the necessary knowledge and context, enabling them to effectively contribute to the project.

\paragraph{Anger:} Anger is characterized by intense feelings of displeasure, hostility, or antagonism toward someone or something perceived as a source of harm or wrongdoing\footnote{\url{https://www.paulekman.com/universal-emotions/what-is-anger/}}. It often manifests in expressions of frustration, outrage, and resentment. Annotators should look for language that conveys aggression, threats, or overt negativity. Examples might include harsh criticism, shouting, or aggressive demands. This emotion is frequently triggered by situations of perceived injustice, insult, or betrayal, and it is crucial for annotators to distinguish it from other negative emotions like disgust or sadness\footnote{\url{https://www.apa.org/topics/anger}}.

\paragraph{Disgust:} Disgust is an emotion that arises from a strong sense of aversion or repulsion toward something offensive, distasteful, or morally objectionable. This feeling can be directed toward people, behaviors, or ideas that violate social norms or personal values. Annotators should identify language that reflects contempt, disdain, or severe disapproval\footnote{\url{https://www.paulekman.com/universal-emotions/what-is-disgust/}}. Common indicators include expressions of revulsion, condemnation, or derogatory remarks. Disgust often accompanies discussions of taboo subjects or unethical actions, requiring careful attention to the context in which these sentiments are expressed.

\paragraph{Surprise:} Surprise is an emotional response to unexpected events or information that deviates from what is anticipated. It can be positive, negative, or neutral, depending on the nature of the unexpected occurrence. Annotators should recognize cues such as exclamations, sudden changes in tone, or language indicating shock or astonishment. This emotion often appears in contexts where new, unforeseen developments are revealed. The intensity of surprise can vary, but key indicators include words and phrases that express sudden realization or discovery\footnote{\url{https://www.paulekman.com/universal-emotions/what-is-surprise/}}. The following annotation guidelines were adhered to consistently throughout the annotation process by our annotators.

\paragraph{Joy:} Joy is a positive emotional state characterized by feelings of happiness, pleasure, and contentment. It often results from fulfilling experiences, successful outcomes, or pleasant interactions. Annotators should identify expressions of delight, gratitude, and satisfaction\footnote{\url{https://www.paulekman.com/universal-emotions/what-is-enjoyment/}}. Language reflecting enthusiasm, laughter, or praise is commonly associated with this emotion. Joy can be expressed in both overt and subtle ways, and it is essential to consider the broader context to accurately capture this emotion in annotations. Annotators should be attentive to both explicit declarations of happiness and more nuanced indications of joy.

\paragraph{Sadness:} Sadness is an emotional response to loss, disappointment, or other unfavorable circumstances. It is characterized by feelings of sorrow, melancholy, and despondency. Annotators should look for language that conveys grief, regret, or a sense of hopelessness\footnote{\url{https://www.paulekman.com/universal-emotions/what-is-sadness/}}. Indicators of sadness include expressions of mourning, lamentation, and heartache. This emotion can be associated with personal experiences, broader societal issues, or empathetic responses to others' suffering. Accurate annotation requires sensitivity to the depth and context of the expressed sadness, ensuring that subtle expressions of this emotion are also captured.

Let us consider some hate speech (HS) and counterspeech (CS) instances to illustrate our annotation process, \\ \\
\noindent \textbf{(HS):} \textbf{The West is heading in the direction of Fascism because Jews are planning to install a world dictator.} \\ \\
\textbf{(CS):} Blaming an entire group for political movements is baseless and harmful. \textbf{(Disgust)}\\

In the above HS and CS example, the counterspeech, it is evident that the response expresses disgust through its explicit rejection of the notion that an entire group can be held responsible for political movements. Phrases such as `baseless' and `harmful` convey strong moral disapproval, which are critical indicators of disgust. The language reflects a profound aversion to the offensive nature of the hate speech, especially in its attempt to generalize blame. By emphasizing the unethical and unfounded nature of these accusations, the counterspeech communicates a sense of repulsion, which supports its classification under the emotion of \textbf{disgust.} Similarly let us consider the following, \\
    
\noindent \textbf{(HS):} \textbf{Muslims are a problem for all other religions.} \\ \\
\textbf{(CS):} I understand that you may have concerns about different religions, but it's important to remember that diversity is what makes our world so beautiful. instead of seeing Muslims as a problem, let's try to celebrate our differences and find common ground. we all share the same desire for love, peace, and understanding. By embracing empathy and affiliation, we can build bridges of dialogue that promote respect and unity among different faiths. let's work together to create a world where everyone feels valued and accepted. \textbf{(Joy)} \\ 

In the analysis of the counterspeech, the response clearly conveys joy by emphasizing the beauty of diversity and the positive potential for unity among different faiths. Language such as ``diversity is what makes our world so beautiful'' and ``we all share the same desire for love, peace, and understanding'' reflect a sense of happiness and appreciation for shared human values. The counterspeech fosters a joyful perspective by encouraging the celebration of differences and the building of connections through empathy. This focus on mutual respect and the vision of a harmonious, inclusive world aligns the response with the emotion of \textbf{joy.}

Following these criteria, annotators meticulously annotated a total of $13,973$ unique counterspeech instances.

\subsection{Advantage of IntentCONANv2}
\label{adav_intconan}
\texttt{IntentCONANv2} represents a significant advancement over its predecessors, including \texttt{IntentCONAN}, by enhancing the quality and structure of counterspeech instances. This dataset builds upon the annotation guidelines established by \citet{gupta-etal-2023-counterspeeches} but focuses on improving content quality by increasing token lengths and ensuring a uniform distribution across four strategies: positive, informative, questioning, and denouncing \citet{hengle2024intentconditioned}. The effectiveness of counterspeech is often linked to its level of detail and comprehensiveness, which can be reflected in its length; a higher token count typically indicates a more thorough and nuanced response, better equipped to address and counteract hate speech. We selected \texttt{IntentCONANv2} for our annotation purposes due to several key advantages. Firstly, it is a large-scale dataset comprising $13,952$ counterspeech instances, offering a substantial foundation for analysis. Secondly, it addresses limitations of earlier datasets, such as \texttt{CONAN} and \texttt{MultiTargetCONAN}, by providing more detailed and informative counterspeech that effectively counters the central aspects of hate speech \citet{hengle2024intentconditioned}. The removal of the humorous strategy is also noteworthy, as it mitigates the risk of subjective or offensive content. Furthermore, \texttt{IntentCONANv2} ensures a consistent representation of counterspeech, with an average of four instances per hate speech example, compared to the two instances in \texttt{IntentCONAN}. Additionally, the dataset emphasizes substantial content, with an average token length of $40.61$, reflecting a focus on creating comprehensive responses that are more effective in countering hate speech.

\subsection{Statistical Analysis on Dataset}
\label{data:statistic}
\begin{figure*}[!t]
  \centering
  \setlength{\belowcaptionskip}{-15pt}
  
  \begin{subfigure}{.45\textwidth}  
    \centering
    \includegraphics[width=0.8\linewidth]{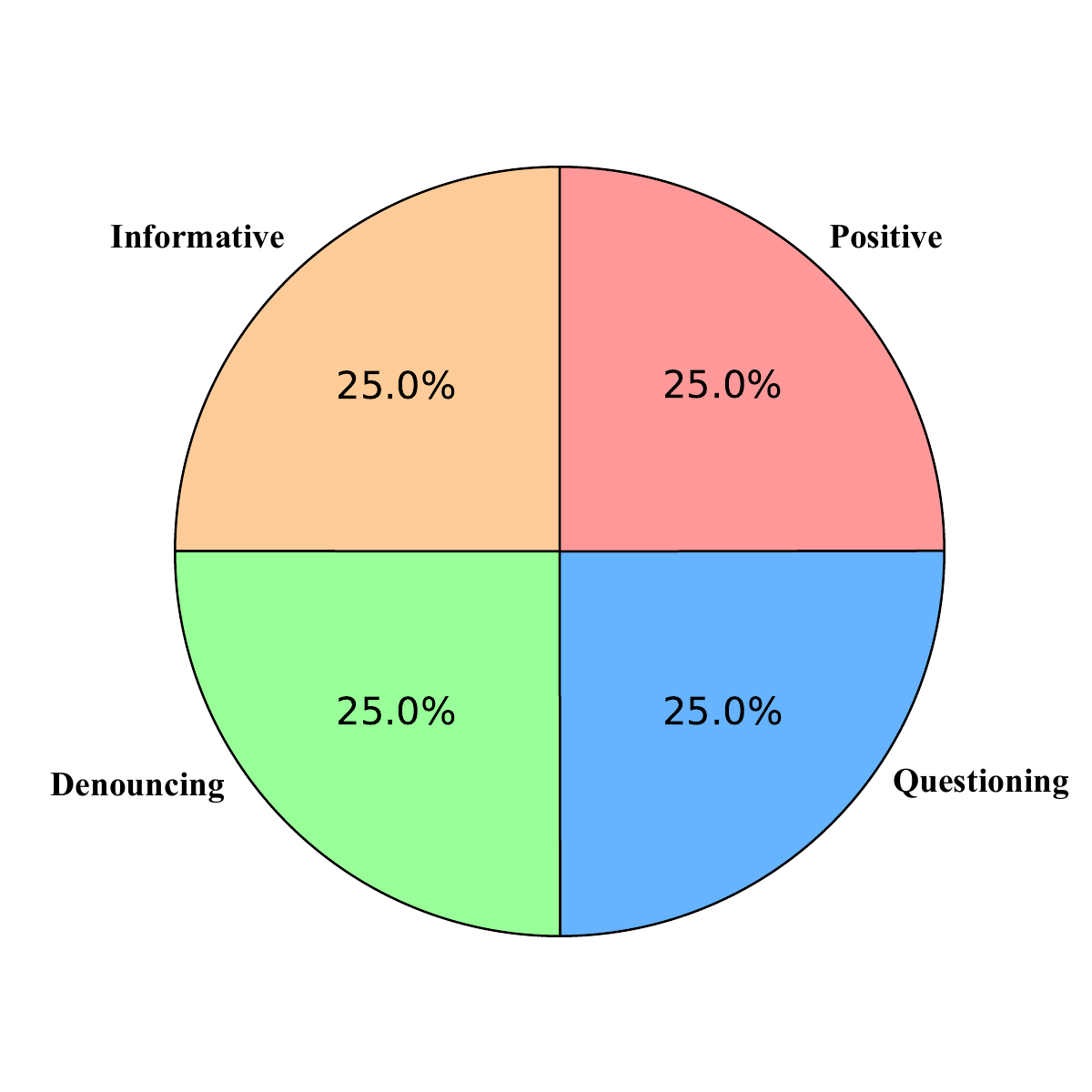}
    \caption{CS, Strategy distribution}
    \label{4_1}
  \end{subfigure}\hfill%
  \begin{subfigure}{.45\textwidth}  
    \centering
    \includegraphics[width=0.8\linewidth]{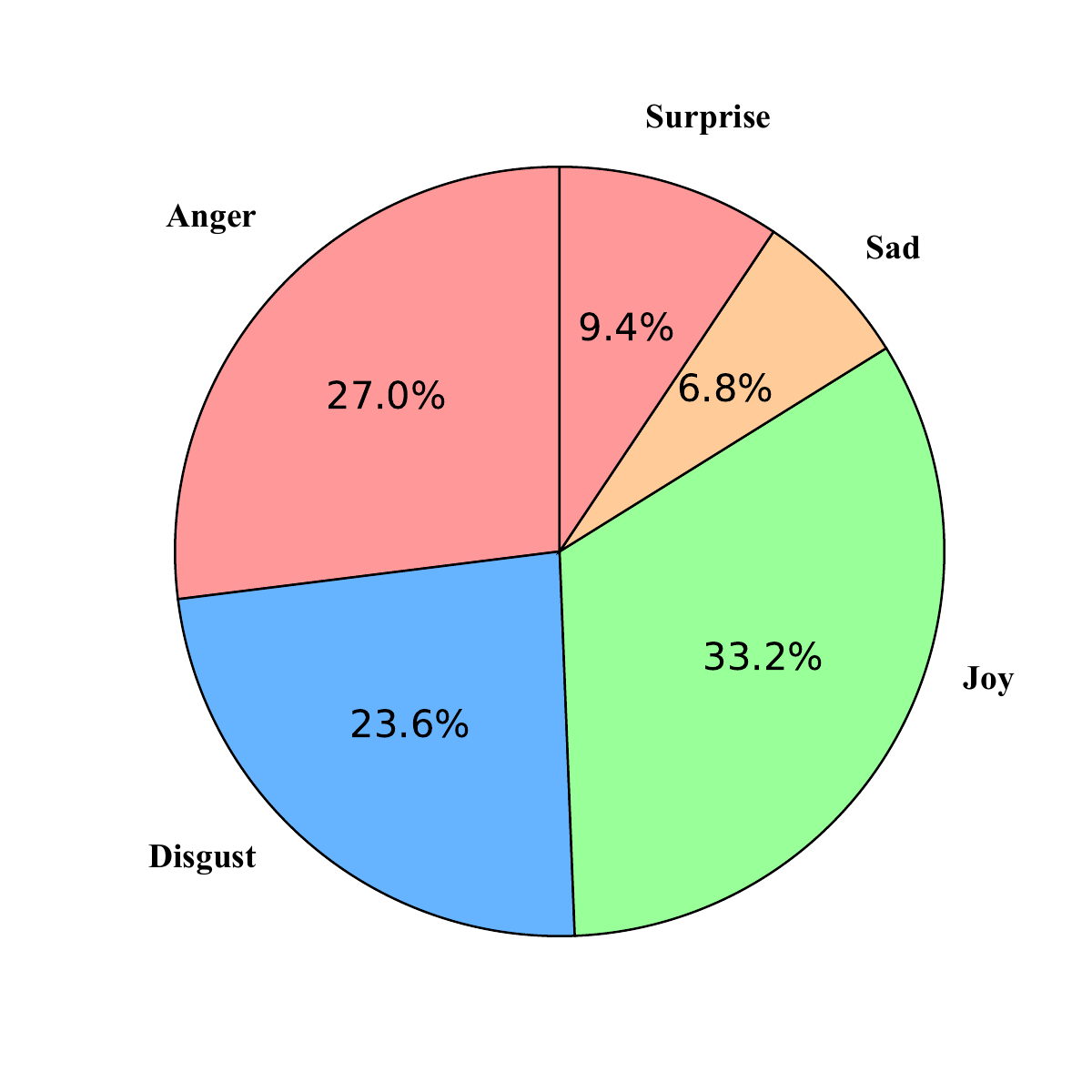}
    \caption{CS, Emotion distribution}
    \label{4_2}
  \end{subfigure}
  
  \medskip
  \medskip
  \medskip
  \begin{subfigure}{.9\textwidth}  
    \centering
    \includegraphics[width= 0.7\linewidth]{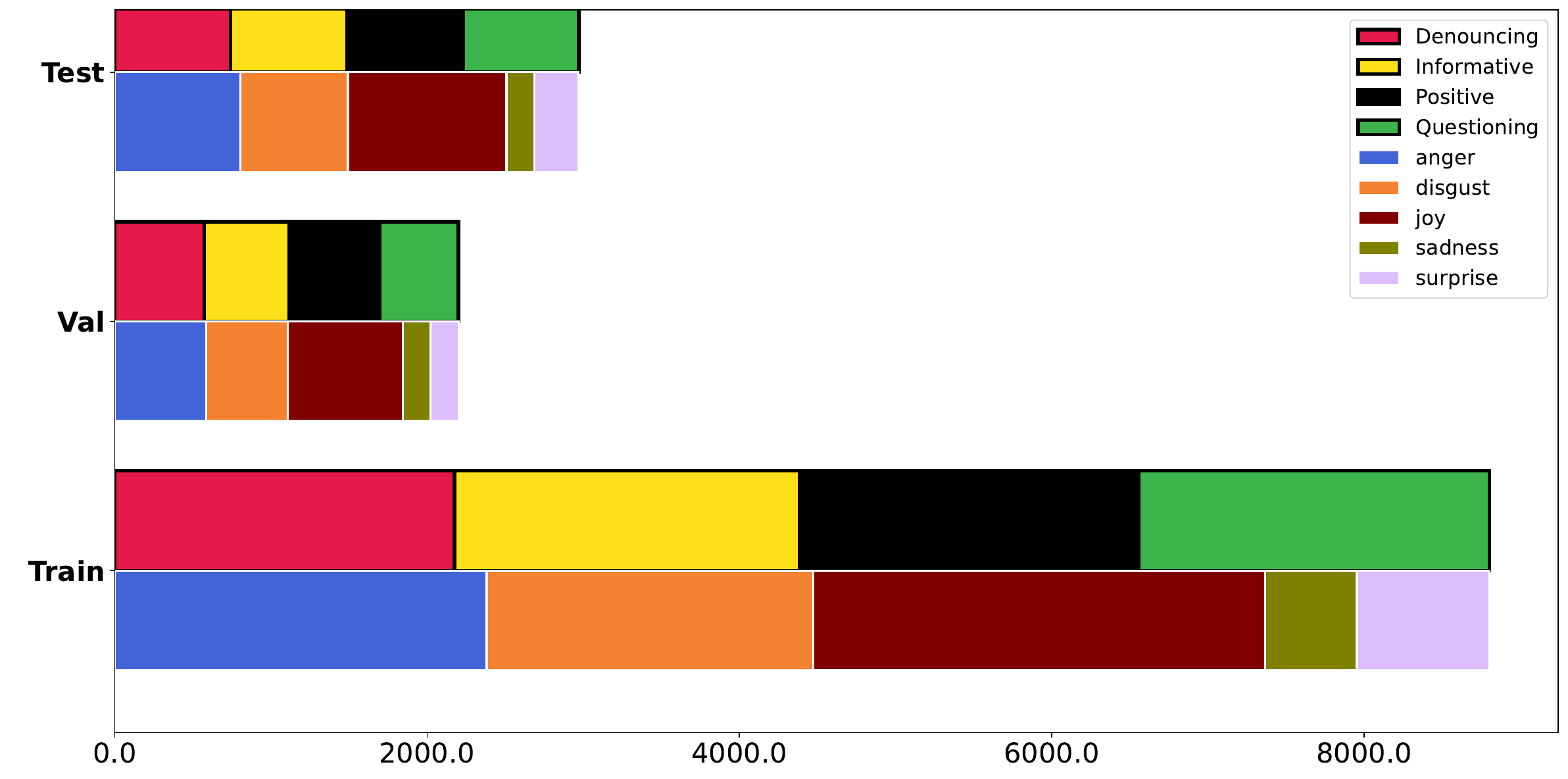}
    \caption{CS strategy and CS emotion distribution across train, val, and test.}
    \label{4_3}
  \end{subfigure}
  
  \medskip
  \medskip
  \begin{subfigure}{.32\textwidth}  
    \centering
    \includegraphics[width= 0.7\linewidth]{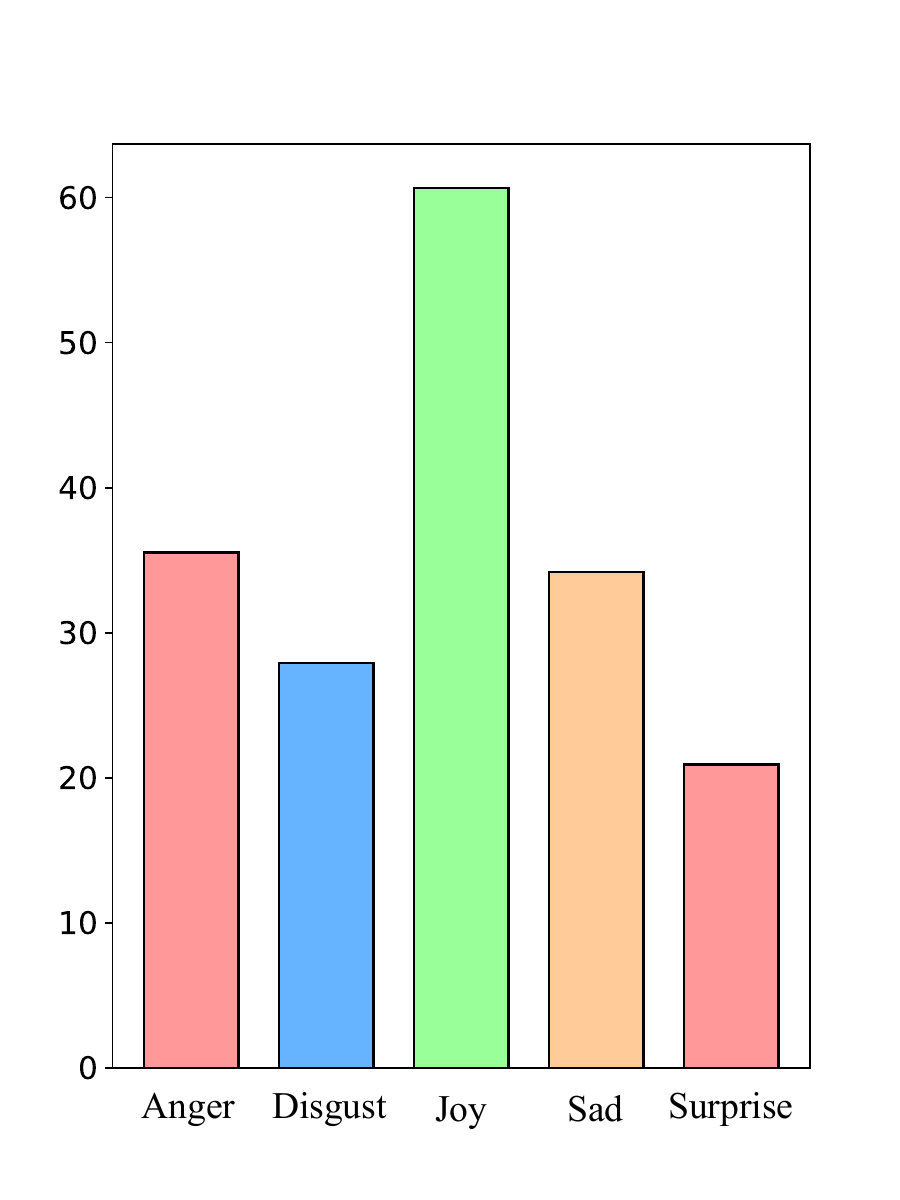}
    \caption{Emotions, mean token length.}
    \label{4_5}
  \end{subfigure}\hfill%
  \begin{subfigure}{.5\textwidth}  
    \centering
    \includegraphics[width=\linewidth]{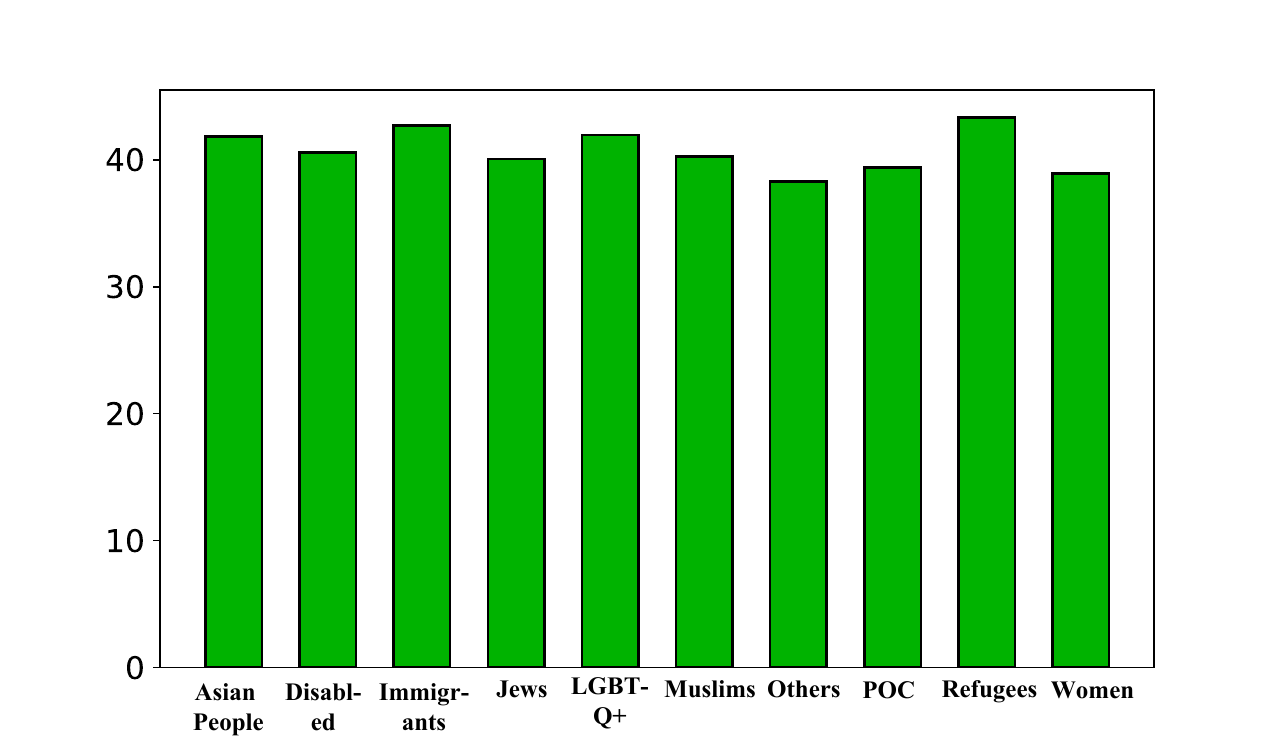}
    \caption{Targets, mean token length.}
    \label{4_6}
  \end{subfigure}
  \medskip
  \medskip
  \caption{Visual exploration of various attribute distribution present in the \texttt{MultiCONAN} dataset.}
  \label{4_all}
\end{figure*}

In the \texttt{multiCONAN} dataset, the distribution of strategy categories across counterspeech is uniform (see Figure \ref{4_1}), with a particular focus on the Emotion category. As illustrated in Figure \ref{4_2}, the distribution of emotion categories within counterspeech instances reveals that the majority fall under the `Joy' category. This predominance signifies the quality and positivity of counterspeech. Joy-based counterspeech typically involves presenting constructive examples and success stories, which necessitate more elaborate responses to effectively build emotional connections through positive narratives. Constructive arguments, in turn, require detailed explanations of alternative viewpoints, further contributing to the need for comprehensive and nuanced responses. Following `Joy,' the categories of `Anger' and `Disgust' are also notable, though to a lesser extent. The `Sad' emotion category has a significantly lower count, indicating its rare occurrence in counterspeech. Figure \ref{4_3} further demonstrates the distribution of both strategy and emotion categories across the training, validation, and testing sets. It is evident that the data splitting is uniform among all categories, ensuring balanced representation and reliable performance assessment across different dataset partitions.

Figure \ref{4_5} presents the mean token length of counterspeech across various emotion categories. The data shows that the `Joy' category has a mean token length of approximately $60$, indicating that counterspeech with joyful emotions tends to be more elaborate. In contrast, other emotion categories, such as `Anger,' `Disgust,' and `Sad,' have a more uniform distribution of mean token lengths, ranging from $25$ to $35$. This variation suggests that counterspeech with more tokens is potentially more effective in targeting and neutralizing hateful comments. Additionally, Figure \ref{4_6} illustrates the mean token length across different target groups, showing a consistent and uniform distribution. This consistency ensures that counterspeech instances are equally detailed and explanatory across all target groups, contributing to the robustness and reliability of the \texttt{multiCONAN} dataset.  

\subsection{Statistical Significance Testing}
\label{stest}

\begin{table}[t!]
    \centering
    \resizebox{\columnwidth}{!}{ 
    \setlength{\tabcolsep}{1mm} 
    \fontsize{9pt}{12pt}\selectfont
    \begin{tabular}{l|c|c|c|c}
        \hline
        \textbf{Metric} & \textbf{T-statistic} & \textbf{$p$-value} & \makecell{\textbf{Significant}} & \makecell{\textbf{Outperform}} \\
        \hline
        \multicolumn{5}{c}{\textbf{Comparison with GPT-3.5 FS}} \\
        \hline
        SC          & 63.34  & 0.0                  & Yes  & Yes  \\
        EC          & 9.31   & $1.71E-20$ & Yes  & Yes  \\
        TC          & 0.96   & 0.33                 & No   & Yes   \\
        BERT Score  & 17.20  & $9.97E-65$ & Yes  & Yes  \\
        METEOR      & -2.37  & 0.018                & Yes  & No   \\
        ROUGE-1          & 8.56   & $1.47E-17$ & Yes  & Yes  \\
        ROUGE-2          & 9.97   & $3.23E-23$ & Yes  & Yes  \\
        ROUGE-L          & 14.21  & $4.57E-45$ & Yes  & Yes  \\
        CoSim  & 0.66   & 0.51                 & No   & Yes   \\
        Toxicity    & 21.30  & $4.85E-97$ & Yes  & Yes  \\
        \hline
        \multicolumn{5}{c}{\textbf{Comparison with GPT-3.5 ZS}} \\
        \hline
        SC          & 38.09   & $2.92E-284$   & Yes   & Yes   \\
        EC          & -22.54   & $4.86E-108$   & Yes   & No    \\
        TC          & -13.22   & $2.50E-39$    & Yes   & No    \\
        BERT Score  & 19.46    & $7.40E-82$    & Yes   & Yes   \\
        METEOR      & -2.43    & $0.015$                  & Yes   & No    \\
        ROUGE-1          & 10.38    & $4.95E-25$    & Yes   & Yes   \\
        ROUGE-2          & 13.77    & $1.64E-42$    & Yes   & Yes   \\
        ROUGE-L          & 15.91    & $7.66E-56$    & Yes   & Yes   \\
        CoSim  & -3.72    & $0.0002$                 & Yes   & No    \\
        Toxicity    & -8.31    &$1.18E-16$     	&Yes	&No	\\ 
	\hline
	\end{tabular}
}
	\caption{Statistical comparison of our model with GPT-3.5 FS and GPT-3.5 ZS across various metrics using T-test and p-values. Positive T-statistic value indicates that our model performs and \( p < 0.05 \) shows the significance.}
	\label{tab:statistical_comparison}
\end{table}

\begin{table*}[!htbp]
\centering

\begin{tabular}{|l|c|c|c|c|c|}
\hline
\textbf{Metric} & \textbf{HIPO\_VT\_3} & \textbf{HIPO\_VT\_5} & \textbf{HIPO\_VT\_7} & \textbf{HIPO\_VT\_10} & \textbf{HIPO\_VT\_3\_DPO} \\
\hline
SC & 0.726 & 0.463 & 0.099 & 0.080 & 1.000 \\
EC & 0.716 & 0.078 & 0.062 & 0.094 & 0.775 \\
TC & 1.000 & 0.009 & 0.315 & 0.000 & 0.966 \\
BERT\_score & 0.878 & 0.358 & 0.806 & 0.158 & 0.923 \\
METEOR & 0.937 & 0.885 & 0.257 & 0.416 & 0.873 \\
ROUGE-1 & 0.816 & 0.647 & 0.933 & 0.535 & 0.941 \\
ROUGE-2 & 0.772 & 0.865 & 0.435 & 0.842 & 0.819 \\
ROUGE-L & 0.777 & 0.643 & 0.877 & 0.595 & 0.868 \\
CoSim & 1.000 & 0.330 & 0.285 & 0.565 & 0.951 \\
Toxicity & 0.929 & 0.137 & 0.418 & 0.098 & 0.995 \\
\hline
\end{tabular}
\caption{Statistical significance across different ablations}
\label{table_1}
\end{table*}

The statistical evaluation of our model against GPT-3.5 Few-Shot (FS) across various metrics highlights significant differences in performance (see Table \ref{tab:statistical_comparison}). For the strategy Conformity (SC) score, our model demonstrates a substantial advantage, with a T-statistic of $63.34$ and a \textit{p}-value of $0.0$, indicating highly significant results. Similarly, for Emotion Conformity (EC), the T-statistic of $9.31$ and a \textit{p}-value of $1.71E-20$ confirm that our model significantly outperforms GPT-3.5 FS. Metrics such as BERT Score, ROUGE-1, ROUGE-2, and ROUGE-L also showcase strong performance by our model, with all \textit{p}-values far below the significance threshold ($p < 0.05$). However, for METEOR, the negative T-statistic ($-2.37$) and a \textit{p}-value of $0.018$ suggest that GPT-3.5 FS slightly outperforms our model in this metric. Interestingly, for Target Conformity (TC) and CoSim, no statistically significant was observed with \textit{p}-values of $0.33$ and $0.51$, respectively. Despite these exceptions, the overall results indicate that our model achieves superior performance across most metrics compared to GPT-3.5 FS.

When compared to GPT-3.5 Zero-Shot (ZS), our model exhibits significant improvements in most metrics, as evidenced by the extremely low \textit{p}-values across strategy Classification (SC), BERT Score, ROUGE metrics (ROUGE-1, ROUGE-2, ROUGE-L), and Toxicity reduction. For instance, SC achieves a T-statistic of $38.09$ with a \textit{p}-value of $2.92E-284$, underscoring the robustness of our model in this task. However, for EC and TC scores, the negative T-statistics ($-22.54$ and $-13.22$) indicate that GPT-3.5 ZS performs better in these areas despite their statistical significance. Additionally, METEOR and CoSim also show slight advantages for GPT-3.5 ZS, with respective T-statistics of $-2.43$ and $-3.72$. Overall, while there are isolated cases where GPT-3.5 ZS performs better or comparably, our model consistently outperforms it across critical metrics such as BERT Score, ROUGE metrics, SC, and Toxicity reduction, demonstrating its effectiveness in counterspeech generation tasks.

Furthermore, we conduct significance testing to compare our model with various ablation configurations (see Table \ref{table_1}). The results indicate no statistical significance in performance, which suggests that our model exhibits robustness to architectural variations and maintains stable performance across diverse configurations \cite{DiLeo2020,Andrade2019, bhojraj2024submental}.

\subsection{ZeroShot Prompt}
\label{ZS}
We used the following prompt for GPT ZeroShot:

\begin{quote}

Generate a counterspeech response to combat hate speech with the following specifications:

\begin{itemize}
    \item \textbf{Intent}: [INTENT]
    \item \textbf{Emotion}: [EMOTION]
\end{itemize}

\textbf{Guidelines}:
\begin{itemize}
    \item Address the hate speech respectfully but firmly
    \item Maintain the specified emotional tone throughout
    \item Focus on constructive dialogue
    \item Avoid toxic or inflammatory language
    \item Align with the specified intent
\end{itemize}

\textbf{Hate Speech}: ``[HATE\_SPEECH]"

\textbf{Counterspeech}:
\end{quote}

\subsection{FewShot Prompt}
\label{FS}
The prompt we used for GPT FewShot is as follows:

\begin{quote}
\textbf{Task}: Generate counterspeech responses to hate speech with specific intent and emotion.

\textbf{Examples}:

\begin{enumerate}
    \item \textbf{Hate Speech}: ``Muslims are terrorists who destroy peace."
    
    \textbf{Intent}: Informative
    
    \textbf{Emotion}: Joy
    
    \textbf{Counterspeech}: ``I'm excited to share that Muslim doctors, teachers, and community leaders contribute positively to our society every day! It's wonderful to see how diverse faiths enrich our communities and promote peace through understanding."

    \item \textbf{Hate Speech}: ``Women don't belong in tech."
    
    \textbf{Intent}: Questioning
    
    \textbf{Emotion}: Surprise
    
    \textbf{Counterspeech}: ``Isn't it fascinating how Ada Lovelace, a woman, wrote the first computer program? And have you seen how mixed-gender tech teams consistently outperform others? These surprising facts challenge our assumptions!"

    \item \textbf{Hate Speech}: ``Immigrants are stealing our jobs."
    
    \textbf{Intent}: Positive
    
    \textbf{Emotion}: Sad
    
    \textbf{Counterspeech}: ``It saddens me to see how economic anxieties can create divisions in our community. The reality is that immigrants actually create millions of new jobs through their entrepreneurship and hard work. Studies show that immigrant-founded businesses employ over 3.7 million Americans. While I understand the fear of job security, I believe we can work together to build an economy that provides opportunities for everyone. Let's focus on supporting each other and creating more opportunities rather than letting fear divide us."
\end{enumerate}

Now generate a counterspeech for:

\textbf{Hate Speech}: ``[HATE\_SPEECH]"

\textbf{Intent}: [INTENT]

\textbf{Emotion}: [EMOTION]

\textbf{Counterspeech}:
\end{quote}

\subsection{Retrival-Based Prompt}
\label{RB}
Generate [INTENT] and [EMOTION] aligned counterspeech addressing these aspects from the hate speech: Hate speech: [HATESPEECH] Relevant examples: [Examples]

\subsection{Computing Information}
\label{Computing Information}
Our research is conducted on the NVIDIA RTX A100 with $80$GB RAM GPU. 

\subsection{Hyper-parameter Information}
\label{Hyper-parameter Information}
Here we have mentioned the hyper-parameters we used for all our experiments,

\begin{itemize}
    \item \text{Batch size}: 4
    \item \text{Learning rate}: 1e-4
    \item \text{Maximum input token}: 512
    \item \text{Maximum output token}: 512
    \item \text{Temperature sampling}: Not used
    \item \text{Max Epoch}: 50
    \item \text{Early stopping used}: Yes
\end{itemize}

\end{document}